\pdfoutput=1

\documentclass[11pt]{article}

\usepackage[final]{acl}

\usepackage{times}
\usepackage{latexsym}
\usepackage{amsmath}
\usepackage{booktabs}
\usepackage{enumitem}

\usepackage{tikz}
\usepackage{tcolorbox}
\usepackage{multirow}

\usepackage{float}
\usepackage{caption}
\usepackage{tabularx}

\usepackage{longtable}
\usepackage{threeparttable}

\usepackage[T1]{fontenc}

\usepackage[utf8]{inputenc}

\usepackage{microtype}

\usepackage{inconsolata}

\usepackage{graphicx}

\title{
Dreaming Up Authors: Factual Prevalence and LLM Hallucinations in Academic Authorship Attribution
}

\author{
  \textbf{Sijia Zhong\textsuperscript{1}},
  \textbf{Tamir Yirga\textsuperscript{2}},
  \textbf{Yijia Ma\textsuperscript{2}},
  \textbf{Thai Le\textsuperscript{3}},
  \textbf{Yifan Hu\textsuperscript{2}}
\\
  \textsuperscript{1}University of Texas at Arlington,
  \textsuperscript{2}Northeastern University,
  \textsuperscript{3}Indiana University
\\
  \small{
    \texttt{qxz9755@mavs.uta.edu},
    \texttt{\{yirga.t, ma.yij, yif.hu\}@northeastern.edu},
    \texttt{tle@iu.edu}
  }
\\
  \small{
    \textbf{Correspondence:} \href{mailto:yif.hu@northeastern.edu}{yif.hu@northeastern.edu}
  }
}

\begin{document}
\maketitle
\begin{abstract}

Large Language Models (LLMs) are widely used for question answering but are prone to hallucinations, yet the underlying causes, particularly the role of factual prevalence in training data, remain underexplored. In this work, we study this through academic authorship attribution, hypothesizing that hallucinations are more likely for less prevalent papers. Using citation count as a proxy for factual prevalence, we curate a dataset of 1.92 million papers across eight scientific disciplines with stratified samples by citation level. Experiments on three state-of-the-art LLMs confirm our hypothesis, revealing a robust inverse relationship between hallucination rate and citation count across all models and disciplines, with hallucination rates exceeding 98\% for rarely cited papers and falling below 37\% for highly cited ones. Our in-depth analysis further shows that these failures primarily reflect retrieval limitations rather than knowledge absence, and hidden-state probing of open-source models shows that factual prevalence is encoded in pre-generation representations, providing convergent mechanistic evidence for the prevalence--hallucination relationship. The code and data are available at \url{https://github.com/quinnzhong1/dreaming-up-authors}.

\end{abstract}

\section{Introduction}

Large language models (LLMs) have demonstrated impressive performance across a wide range of natural language processing tasks and are increasingly adopted in everyday applications, business operations, and scientific research.
Despite these advances, LLMs are notorious for their hallucinations, generating plausible but incorrect or misleading responses. Such errors often include instruction inconsistency, logical inconsistency, factual contradiction, or fabrication~\cite{Huang_2025_hallucination_survey}.
A particularly concerning type of hallucination is fact-conflicting hallucination~\citep{zhang2025sirenssongaiocean}, in which generated content contradicts established world knowledge, such as producing fake references or incorrect author names for a research paper. These failures pose serious risks in knowledge-intensive domains like scientific research, such as the case of hallucinated citations~\cite{naddaf2026hallucinated}, highlighting the need to better understand the underlying causes of hallucination.

In this work, we investigate \emph{\textbf{factual prevalence}} as a potential contributor to hallucinations. Factual prevalence refers to how frequently a specific fact, such as a name, event, or concept, appears in an LLM's training data. Limited exposure to certain information may explain why LLMs exhibit weaker performance or hallucinate in less-represented domains~\citep{mallen2023_whenNotToTrustLMs}. Previous studies have shown that LLMs can memorize factual knowledge~\cite{chowdhery2022palm}, and that this memorization improves with model scale~\cite{carlini2022quantifying}, but struggles with long-tail knowledge~\cite{Kandpal2022_LLMStruggleLongTail,sun2024headtotail} (full related work is provided in Section~\ref{sec:appendix_related_work}).
However, systematic evaluation of this relationship, specifically, how the prevalence of the required knowledge in training data affects hallucinations, remains largely unexplored.
A major obstacle is that the data used to pre-train commercial state-of-the-art (SOTA) LLMs is not available to the academic community, making direct investigation infeasible.

To address this challenge, we adopt \emph{academic authorship attribution}, an underexplored task in LLM hallucination research, and propose citation counts as a practical proxy for factual prevalence.
We focus on the following research question (RQ): {\em For queries involving authorship information, how does factual prevalence, approximated by a paper’s citation count, affect the likelihood of hallucinations in LLM-generated responses?}

Academic papers are well indexed and broadly accessible through online databases, and citation counts naturally reflect their visibility. Given that SOTA LLMs utilize much knowledge from the Internet, which should include those academic resources, we accordingly adopt citation count as a practical proxy for factual prevalence and study a concrete, measurable question: \emph{Does the hallucination rate of LLMs decrease as a paper's citation count increases?} In answering such a question, we hypothesize that hallucination rates are inversely related to factual prevalence: LLMs are more likely to hallucinate when attributing authors of less prevalent, low-citation papers. This hypothesis aligns with the training dynamics of LLMs, which learn next-token probabilities from large-scale corpora, with frequently occurring facts reinforced, while rare facts are more susceptible to mis-prediction and hallucination.

To investigate this hypothesis, we curated a high-quality dataset of 1.92 million academic papers spanning eight scientific disciplines: Medicine, Biology, Chemistry, Computer Science, Psychology, Physics, Materials Science, and Mathematics.
For efficient evaluation of LLMs, we further sampled a dataset of 9,108 papers stratified across disciplines and citation bins. We then verified that citation counts correlate with factual prevalence by comparing them to Google search hits and the frequency of paper titles in the Infini-gram corpus~\cite{Liu2024InfiniGram}. Next, we tested three SOTA LLMs with carefully designed prompts for authorship prediction. Our analysis
demonstrated a consistent negative correlation between hallucination rates and citation counts across models and disciplines.

\textbf{Our main contributions are:}

\begin{itemize}[
   leftmargin=1em,
    itemsep=0.2em,
    topsep=0.2em,
    partopsep=0em,
    parsep=0em
]
    \item
    We curated 1.92M papers with associated metadata, and a 9,108-paper evaluation subset stratified across eight disciplines and 13 citation bins.
    We will release this dataset to support future research in hallucination detection.

    \item
    We established that citation count positively correlates with both Google search hits and $n$-gram frequency across two large LLM pretraining corpora, justifying the use of citation count as a proxy for factual prevalence.

    \item
    We demonstrate a robust inverse relationship between citation count and hallucination rate across three SOTA LLMs and eight disciplines.

    \item We demonstrate via a multiple-choice recognition experiment that LLM recall failures for low-prevalence papers primarily reflect retrieval limitations rather than knowledge absence.

    \item We provide hidden-state evidence from open-source models that LLMs encode paper prevalence in pre-generation representations, linking the
    prevalence--hallucination relationship to the model's internal state before generation.

\end{itemize}

\section{Academic Authorship Dataset}
\label{sec:dataset}

To investigate the relationship between citation count and hallucination rate in authorship attribution, we constructed a large-scale academic dataset.
Our guiding curation principle was that the papers must constitute a true
random sample published before the release dates of the LLMs we evaluated. The only source that we are aware of that enables such sampling is the Microsoft Open Academic Graph (OAG v2)~\cite{OAGv2}. Since the OAG v2 snapshot dates to 2019 with outdated citation counts, we supplemented metadata using the Semantic Scholar API.

\textbf{Microsoft Open Academic Graph.} With its hundreds of millions of papers spanning diverse disciplines, OAG provides a broadly representative snapshot of the academic literature. We randomly sampled 10 million papers, using the metadata available in OAG, including titles, authors, publication years, and DOIs.

\textbf{Semantic Scholar API.} We enriched OAG records by matching with Semantic Scholar to obtain up-to-date citation counts, canonicalized author names, and field classifications (as of September 2025). Note that Semantic Scholar does not offer an API for random sampling of papers, necessitating our two-step approach.

\subsection{Quality Assurance Pipeline}

To ensure data quality, we applied a rigorous multi-stage filtering process (Figure~\ref{fig:data_filtering}), retaining 1,921,209 papers from 9,999,863 initial OAG records (19.2\% retention rate).
The filters (with \% removed) are as follows.

\textbf{Document type (56.9\%):} Retained only journal and conference papers to focus on standard research articles, excluding books, patents, and book chapters.
\textbf{Author Count (0.3\%):} Removed papers with zero or 20+ authors.
\textbf{Language (47.9\%):} Applied FastText language detection to retain English papers (confidence $\geq$ 0.8).
\textbf{Author Name Patterns (0.8\%):} Removed papers with problematic author name formats such as institutional names, single-token names, and anomalies indicating encoding errors.
\textbf{Title/Author Match (13.5\%):} Verified exact title and first author last name matches between OAG and Semantic Scholar to eliminate false positives from Semantic Scholar title-based search.
The complete filtering specifications are given in  Appendix~\ref{sec:appendix_filtering}.

\subsection{Dataset Characteristics}

Our final dataset of 1,921,209 papers exhibits diverse characteristics that enable comprehensive hallucination analysis. We present key distributions below; comprehensive statistics
are given in Appendix~\ref{sec:appendix_stats}.

\textbf{Citation Distribution.}
Citations follow a heavily right-skewed distribution
(Table~\ref{tab:citations}): median = 7, with 36.9\% of papers having 0--2 citations and 78.6\% fewer than 33. This skewness motivates our exponential binning in stratified sampling, providing fine granularity at low-citation ranges while maintaining sufficient representation at high-citation ranges.

\setlength{\tabcolsep}{4pt}
\begin{table}[tb!]
  \centering
  \captionsetup{font=small}
  \footnotesize
  \begin{tabular}{@{}lrrr@{}}
    \toprule
    \textbf{Citation Range} & \textbf{Count} & \% & \textbf{Cumulative} \% \\
    \midrule
    0--2 & 708,267 & 36.9 & 36.9 \\
    3--4 & 137,934 & 7.2 & 44.1 \\
    5--8 & 185,875 & 9.7 & 53.7 \\
    9--16 & 231,214 & 12.0 & 65.8 \\
    17--32 & 247,346 & 12.9 & 78.6 \\
    33--64 & 204,806 & 10.7 & 89.3 \\
    65--128 & 123,049 & 6.4 & 95.7 \\
    129--256 & 54,789 & 2.9 & 98.5 \\
    257--512 & 19,228 & 1.0 & 99.5 \\
    513--1024 & 6,232 & 0.3 & 99.9 \\
    1025--2048 & 1,783 & 0.1 & 100.0 \\
    2049--4096 & 508 & 0.0 & 100.0 \\
    4097+ & 178 & 0.0 & 100.0 \\
    \midrule
    \textbf{Total} & \textbf{1,921,209} & \textbf{100.0} & \\
    \bottomrule
  \end{tabular}
    \caption{Citation distribution across the full dataset.}
      \label{tab:citations}
\end{table}

\textbf{Field Distribution.} The dataset covers 19 academic disciplines.
We focus on eight academic fields: Medicine, Chemistry, Biology, Computer Science, Materials Science, Physics, Psychology, and Mathematics. These fields are identified based on the field-of-study distribution of a random sample of 100,000 papers drawn from the initial data pool prior to filtering (Table~\ref{tab:fos-dist-sampled}). These fields account for 76.95\% of papers in the final curated dataset and provide sufficient sample sizes across citation bins to support robust analysis (Table~\ref{tab:fos-dist}).

\textbf{Author Count and Temporal Coverage.} Papers contain 1--20 authors (mean = 3.2, median = 2), with 31.5\% being single-authored and a long tail extending to 20-author collaborations
(Table~\ref{tab:authors-dist}).
Author count varies substantially by field, reflecting different collaboration norms (Table~\ref{tab:authors-by-field-dist}): Medicine and Biology both average 4.1 authors (large collaborative teams) while Mathematics averages 2.1 authors (smaller theoretical teams). Publications span from 1800 to 2019 (Table~\ref{tab:temporal-dist}), with 68.2\% published between 2000--2019, reflecting the exponential growth of academic publishing in the digital era.

\textbf{Field-Specific Characteristics and Research Implications.} The cross-field variation in authorship and citation patterns provides natural experimental controls for our hallucination analysis. Citation counts vary substantially across disciplines (Table~\ref{tab:citations-by-field-dist}): Biology averages 45.0 citations while Mathematics averages 25.4 citations, tracking with their mean author counts. However, other fields decouple these factors: Psychology averages 39.2 citations with 2.4 authors, while Materials Science averages 26.0 citations with 3.8 authors. This variation allows us to test whether hallucination patterns are driven primarily by citation prevalence or confounded by factors such as list length or field-specific conventions.

\section{Citation Count as a Proxy for Factual Prevalence}

In the context of academic publications, we use citation count as a proxy for academic visibility or salience, and assume that it correlates with factual prevalence in LLM training data.

Ideally, factual prevalence would be measured directly from the frequency of specific facts in the actual training corpora of LLMs. However, the training data used by most state-of-the-art LLMs remains proprietary and undisclosed. Although Common Crawl~\cite{commoncrawl} is known to contribute to many LLM training pipelines, it is difficult to analyze reliably: the raw data is vast, noisy, and heavily filtered before it is used for training~\cite{brown2020gpt3}. As a result, the effective data seen by LLMs differs substantially from raw Common Crawl and is opaque to external researchers.

To validate that citation count is a good proxy for factual prevalence, we test the hypothesis that LLM training data contains more occurrences of highly cited papers than less cited ones. We adopt two complementary analysis approaches: \textbf{Google Search Hits Analysis} and \textbf{LLM Corpus Occurrence Analysis}.

To support these analyses, we perform stratified random sampling and select 6,048 papers from the full 1.92M dataset, targeting approximately 500 papers per citation bucket, with fewer samples in the highest citation buckets due to data availability.

\begin{figure}[t]
  \centering
  \includegraphics[width=\columnwidth]{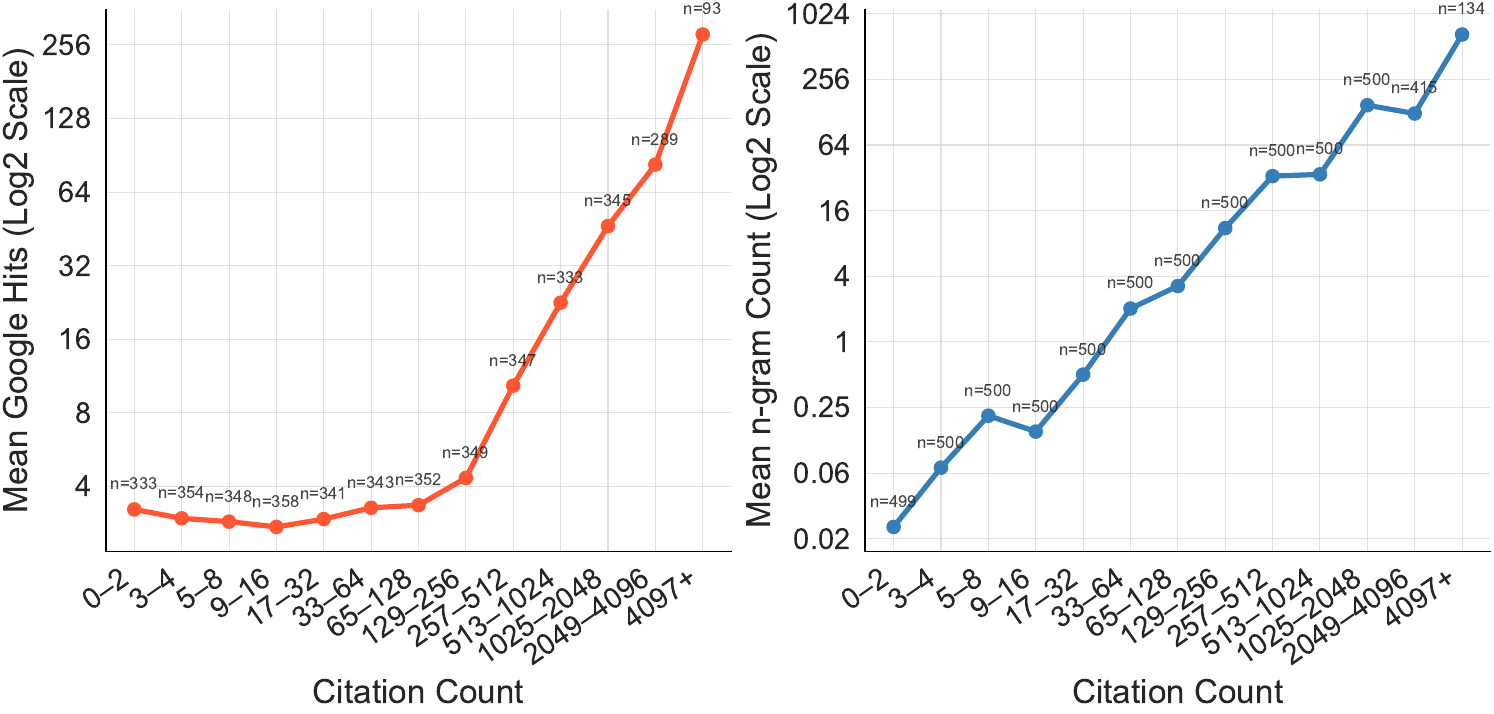}
  \captionsetup{font=small}
    \caption{
    Proxy signals for factual prevalence vs citation bucket.
    Left: Log-Scale mean Google search hits (after outlier filtering).
    Right: Log-Scale mean Infini-gram $n$-gram counts in the OLMo2-32B corpus.
    }
  \label{fig:citation_count_proxy}
\end{figure}

\subsection{Google Search Hits Analysis}

The number of results returned by a web search can serve as a rough indicator of the prevalence of queried content on the web. Motivated by this observation, and noting that common crawled web data constitutes a major component of many LLM pretraining corpora~\cite{brown2020gpt3}, we examine whether highly cited papers tend to exhibit greater web presence. To this end, we collect Google search hit counts for sampled papers using their titles and author lists as query strings.

For each paper, we construct a query by concatenating its title and author names, enclosing each component in quotation marks (e.g., \texttt{``Paper Title'' ``Author1'' ``Author2'' ``Author3''}), and retrieve the corresponding number of hits using BrightData's Google Search API.

We observe substantial outliers, particularly in lower citation buckets. In the 0–2 bucket, papers with zero citations can exhibit extremely high hit counts due to short titles that frequently match unrelated web content. For example, the query \texttt{``Is this tragic or comic'' ``H.~White''} yields 49{,}300{,}000 hits despite having zero citations; its query string length (31) is far shorter than the full-dataset average (129.97), increasing the likelihood of spurious matches.

To reduce the impact of such outliers, we remove papers whose hit counts exceed the 80th percentile within each citation bucket (details in Appendix~\ref{sec:appendix_web_presence}). The filtered dataset contains 4,185 papers, with most buckets retaining over 300 samples. Summary statistics are reported in Figure~\ref {fig:citation_count_proxy}~(left).
It is clear that the mean hit counts increase with citation level,
supporting the hypothesis that \emph{highly cited papers tend to appear more frequently in web-crawled data}.

\subsection{LLM Corpus Occurrence Analysis}

To validate the relationship between citation count and frequency of appearance in large LLM training corpora, we leverage Infini-gram~\cite{Liu2024InfiniGram}, a large-scale $n$-gram database derived from trillion-token corpora that provides sentence-level occurrence counts. For each paper, we construct a canonical query by concatenating its title with the first author's name, using only the first author to reduce noise from varying author order and formatting. Following the API documentation, we join the capitalized title and first author with the logical operator AND (e.g., \texttt{Attention is all you need AND Ashish~Vaswani}).

We query the Infini-gram API over the full sampled dataset of 6,048 papers and record the corresponding $n$-gram counts. This procedure is applied to two corpora: OLMo2-32B-Instruct (6.1T tokens), a publicly available instruction-tuning corpus built on OLMo's openly documented pretraining data, and RedPajama (1.4T tokens), a public reproduction of LLaMA's pretraining dataset. Detailed results are reported in Appendix~\ref{sec:appendix_corpus_occurrence}.
Results on OLMo2-32B corpus are summarized in Figure~\ref{fig:citation_count_proxy}~(right).

Across both corpora, the mean $n$-gram count exhibits an overall increasing trend with citation bucket, despite minor local fluctuations. For the lowest citation buckets (0–2, 3–4), mean counts remain near zero, indicating that low-citation papers rarely appear in these corpora, whereas higher citation buckets, particularly above 128, show consistently higher mean counts. Overall, these results suggest that \emph{highly cited papers are more reliably represented in open LLM pretraining datasets}.

\subsection{Why Citation Count Serves as a More Reliable Prevalence Proxy}

While both analyses support our hypothesis, neither is suitable as a standalone proxy: Google hit counts are unstable due to unrelated web matches and query-length sensitivity, and Infini-gram corpus counts are highly sensitive to minor formatting variations such as capitalization and punctuation (see Appendix~\ref{sec:appendix_web_presence} and \ref{sec:appendix_corpus_occurrence}). In contrast, citation counts are easy to obtain, stable, and consistently defined, making them a more reliable and practical proxy for factual prevalence.

\section{Experiment Setup}
\label{sec:methodology}

\subsection{Experimental Sampling Strategy}

The highly right-skewed, long-tailed distribution of citation counts, together with the substantial cost and time required for LLM querying, makes it neither feasible nor practical to evaluate the full set of 1.92 million papers. We therefore adopt a stratified sampling strategy, sampling up to 100 papers per field per citation bin across the eight fields (covering 77\% of the full dataset) and thirteen exponentially spaced citation bins.
Because the 2,049–4,096 citation bin contains only 508 papers (avg 64 per field), and the 4,097+ bin contains only 178 papers (avg 22 per field), our sampling target of 100 papers per (field, citation bin) stratum could not always be met. The final sample consists of 9,108 papers (vs theoretical maximum of 10,400), with shortfalls primarily in high-citation bins. Details on the sampling procedure and rationale for exponential binning are provided in Appendix~\ref{sec:appendix_data_sampling}.

This sampling strategy provides balanced coverage across the citation spectrum while maintaining sufficient statistical power for analyzing correlations between citation counts and hallucination rates. Table~\ref{tab:field_sample_sizes} presents the actual distribution of papers across fields. To verify that our sample size of 100 papers per bin provided stable hallucination rate estimates, we computed standard deviations and 95\% confidence intervals for HR$_2$ (Equation~\ref{eqn:hr2}) within each  stratum. Across all 104 populated strata, the mean standard deviation was 0.197, with 95\% confidence intervals averaging $\pm$0.065. Results in Section~\ref{sec:results} will show that this stratification ensures sufficient statistical power for detecting the correlation patterns.

\setlength{\tabcolsep}{1pt}
\begin{table}[tb!]
  \centering
  \footnotesize
  \captionsetup{font=small}
  \begin{tabular}{@{}lrrr@{}}
    \toprule
    \textbf{Field} & \textbf{Papers} & \textbf{Avg Authors} & \textbf{Avg Citations} \\
    \midrule
    Medicine & 1,230 & 5.0 & 659 \\
    Biology & 1,179 & 4.6 & 506 \\
    Computer Sci. & 1,174 & 3.2 & 601 \\
    Psychology & 1,154 & 2.9 & 413 \\
    Chemistry & 1,134 & 3.8 & 356 \\
    Mathematics & 1,088 & 2.2 & 355 \\
    Physics & 1,082 & 3.7 & 297 \\
    Materials Sci. & 1,067 & 4.2 & 266 \\
    \midrule
    \textbf{Total} & \textbf{9,108} & \textbf{3.7} & \textbf{438} \\
    \bottomrule
  \end{tabular}
    \caption{Stratified sampling statistics per field.}
  \label{tab:field_sample_sizes}
\end{table}

\subsection{LLM Prompt Design for Academic Authorship Attribution}
\label{sec:prompt_design}

To elicit consistent and as accurate as possible author attributions from LLMs, we design a structured few-shot prompt template that enforces strict formatting constraints (e.g., following Western name order, returning only author names, and excluding any additional explanations), while providing multiple examples for reference.

The template takes as input a paper's title and publication year, and uses the same question format to construct an academic authorship attribution query: \texttt{Who are the authors of the paper titled '\{title\}' which was published in \{year\}?}
The model is instructed to return a comma-separated list of author names in the specified format.
The complete verbatim prompt template is provided in Appendix~\ref{sec:appendix_prompt_design}.

\subsection{LLMs Tested}
\label{sec:llm_config}

We evaluated three state-of-the-art large language models to assess the generalizability of our findings. \textbf{GPT-4o} is used as our primary model due to its strong performance on knowledge-intensive tasks and widespread adoption. \textbf{DeepSeek-R1} represents a reasoning-oriented model that employs explicit chain-of-thought inference. \textbf{Claude Sonnet 4.5} represents a closed-source model with internal reasoning capabilities, providing a contrast to models with more explicit reasoning traces.

\subsection{Hallucination Rate Definitions}
\label{sec:hr_definitions}

Since the average paper has 3.7 authors, a binary label for evaluating LLM hallucinations would unfairly penalize partially correct answers. We therefore propose the following more nuanced metrics. For each paper, we compute three quantities (Figure~\ref{fig:venn_example}) by comparing the ground-truth author set with the LLM-generated set:

\begin{figure}[tb!]
\centering
\begin{tikzpicture}[every node/.style={font=\sffamily\bfseries}, scale=0.72, transform shape]

\coordinate (LeftCircle) at (-2, 0);
\coordinate (RightCircle) at (2, 0);

\def\xrad{3.3cm}
\def\yrad{1.3cm}

\definecolor{ColorLeft}{HTML}{fbb4ae}
\definecolor{ColorRight}{HTML}{b3cde3}

\fill[ColorLeft, fill opacity=0.5] (LeftCircle)
    ellipse [x radius=\xrad, y radius=\yrad];
\fill[ColorRight, fill opacity=0.5] (RightCircle)
    ellipse [x radius=\xrad, y radius=\yrad];

\draw[ColorLeft, thick] (LeftCircle)
    ellipse [x radius=\xrad, y radius=\yrad];
\draw[ColorRight, thick] (RightCircle)
    ellipse [x radius=\xrad, y radius=\yrad];

\node at (-2.2, 1.5) [black] {Ground Truth ($M+U$)};
\node at (2.6, 1.5) [black] {LLM Generated ($M+H$)};

\node at (-3.2, 0) [align=center] {
    \textcolor{black!70!black}{U}\\
    \textcolor{black!70!black}{Unretrieved}\\
    \small\textit{\textcolor{black!50!white}{B. Jones}}\\
    \scriptsize \textcolor{black!50!white}{(In GT, not LLM)}
};

\node at (0, 0) [align=center] {
    M\\
    Matches\\
    \small\textit{\textcolor{black!50!white}{A. Smith}}\\
};

\node at (3.2, 0) [align=center] {
    \textcolor{black!70!black}{H}\\
    \textcolor{black!70!black}{Hallucinations}\\
    \small\textit{\textcolor{black!50!white}{C. Brown}}\\
    \scriptsize \textcolor{black!50!white}{(In LLM, not GT)}
};

\end{tikzpicture}

\captionsetup{font=small}
\caption{Components in the hallucination rates.}
\label{fig:venn_example}
\end{figure}
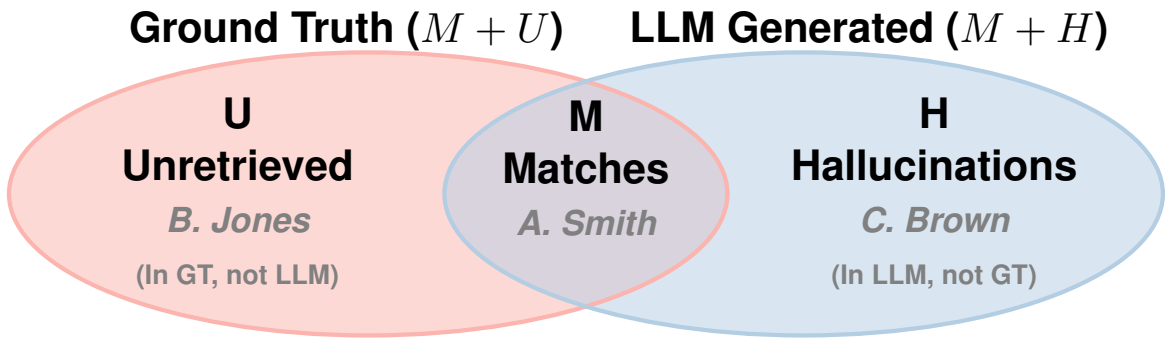

\begin{itemize}[
   leftmargin=1em,
    itemsep=0.2em,
    topsep=0.2em,
    partopsep=0em,
    parsep=0em
]
    \item $\mathbf{M}$ ({Matches}): authors appearing in both lists (true positives)
    \item $\mathbf{H}$ ({Hallucinations}): authors generated by the LLM but absent from ground truth (false positives)
    \item $\mathbf{U}$ ({Unretrieved}): authors in ground truth but not generated by the LLM (false negatives)
\end{itemize}

We define three distinct hallucination rate metrics to quantify the degree of hallucination in LLM-generated author list responses for each paper.
All of them range from 0 (no hallucination) to 1 (pure hallucination):

\textbf{Max-Normalized Error Rate (HR$_1$)} measures error relative to the larger of the two lists, making this metric robust to extreme generation lengths:

\begin{equation}
\text{HR}_1 = 1 - \frac{|M|}{\max(|M|+|H|, |M|+|U|)}
\end{equation}

\textbf{Jaccard Hallucination Rate (HR$_2$)} measures the proportion of errors within the total pool of unique names. It is mathematically equivalent to 1 minus the Jaccard similarity and is a symmetric, length-invariant measure of overall prediction error:

\begin{equation}
\text{HR}_2 = \frac{|H|+|U|}{|M|+|H|+|U|}\label{eqn:hr2}
\end{equation}

\textbf{Harmonic Hallucination Rate (HR$_3$)} penalizes cases where hallucinated and missing authors co-occur, emphasizing severe failures in which an LLM simultaneously invents and omits authors, analogous to the reverse of an F1 measure.

\begin{equation}
\text{HR}_3 = \frac{2 |H| |U|}{(|M|+|H|) |U| + (|M|+|U|) |H|}
\end{equation}

We found that all three metrics yield directionally consistent results and are highly correlated (Pearson correlations $\ge$ 0.98, Table~\ref{tab:metric_correlations}).  \textbf{We report HR$_2$ as the primary metric} because it balances both error modes: fabricated authors and omitted true authors. We additionally evaluated HR$_4$ (Hallucination-Only Rate: the fraction of predicted names that are fabricated) and HR$_5$ (Omission-Only Rate: the fraction of true authors the model failed to retrieve); all five metrics exhibit the same ordinal trends across citation bins, with HR$_4$ and HR$_5$ differing by only 0.007 on average (maximum 0.022). Additional analyses of alternative metrics and the separate fabrication/omission components are provided in Appendix~\ref{sec:appendix_HR_definitions}.

\subsection{Evaluation Pipeline}
\label{sec:eval_comparison}

We implemented a hard-coded evaluation pipeline using deterministic algorithmic fuzzy string matching to compare LLM-generated author names against ground truth. The evaluation employs a multi-stage matching process that accounts for common formatting variations (e.g., ``J. Smith'' vs. ``John Smith''), handles special characters and diacritics, and manages name order differences. To establish a match, both the last name and first initial must match after normalization. Based on this process, the algorithm outputs the three quantities (M, H, U) for each LLM-generated answer.

For comparison, we also experimented with LLM self-evaluation. A detailed comparison between LLM-based and hard-coded evaluation methods is given in Appendix~\ref{sec:appendix_eval_comparison}. We found that HR$_2$ given by the two approaches differs minimally (with RMSE of 0.067, and 96.9\% of the time within 0.1). We adopt \textbf{hard-coded evaluation} as our primary evaluation method due to its superior reproducibility, lower cost, and independence from the specific LLM used for evaluation.

\section{Experiment Results}
\label{sec:results}

\begin{figure}[h]
  \centering
  \includegraphics[width=\columnwidth]{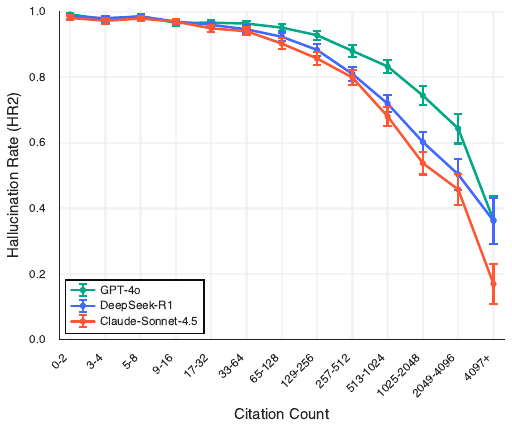}
  \captionsetup{font=small}
\caption{Hallucination rate (HR$_2$) decreases monotonically with citation count across models.}
  \label{fig:hr_by_citations}
\end{figure}

We plot the hallucination rate (HR$_2$) as a function of citation count in Figure~\ref{fig:hr_by_citations}. A consistent pattern is seen across all three LLMs: {\em hallucination rates
decline substantially as citation count increases}. For papers with 0–2 citations, HR$_2$ exceeds 0.98 across all models, indicating near-complete failure to identify authors. Performance improves monotonically with citation count, reaching HR$_2$ values of 0.364 (GPT-4o), 0.362 (DeepSeek-R1), and 0.169 (Claude Sonnet 4.5) for the highest-citation papers.
This suggests a fundamental relationship between a paper's visibility
and the model's ability to recall its authors.
The error bars in Figure~\ref{fig:hr_by_citations} represent 95\% confidence intervals, further validating that this relationship is statistically significant.

Across all fields and models, citation count exhibits a statistically significant negative correlation with hallucination rate. As shown in Table~\ref{tab:spearman_all_models}, this relationship holds consistently across all 24 field--model combinations. The strength of the correlation varies by field. Computer Science, Medicine, and Psychology show stronger negative correlations, while other fields also show statistically significant negative correlations. Among the three LLMs, DeepSeek and Claude exhibit stronger negative correlations than GPT-4o.

\begin{figure}[tb!]
  \centering
  \includegraphics[width=\columnwidth]{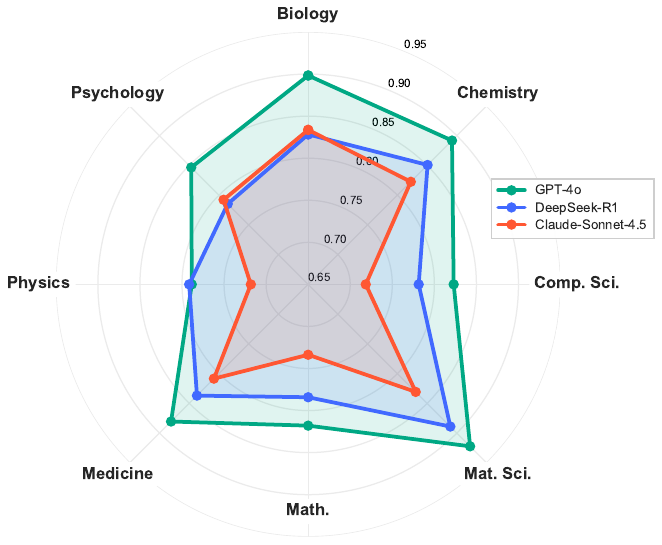}
  \captionsetup{font=small}
  \caption{Field-specific hallucination rate (HR$_2$) distributions across three LLMs.}%
  \label{fig:field_variance}
\end{figure}

To further confirm the relative performance of LLMs, Figure~\ref{fig:field_variance} compares the overall HR$_2$ distributions across fields for the three models. All models rank fields similarly: Materials Science and Chemistry consistently exhibit the highest hallucination rates, whereas Computer Science and Physics show the lowest. This cross-model agreement suggests that field-level difficulty is driven primarily by intrinsic domain characteristics, such as naming conventions, citation structure, and training data coverage, rather than model-specific effects. Among the models, Claude Sonnet exhibits the lowest overall hallucination rates, followed by DeepSeek-R1, with GPT-4o performing worst.

\setlength{\tabcolsep}{5pt}
\begin{table}[tb!]
\centering
  \captionsetup{font=small}
  \footnotesize

  \begin{tabular}{@{}lccc@{}}
    \toprule
    \textbf{Field} & \textbf{GPT-4o} & \textbf{DeepSeek} & \textbf{Claude} \\
    \midrule
    Biology & -0.400 & -0.489 & -0.420 \\
    Chemistry & -0.323 & -0.406 & -0.383 \\
    Computer Sci. & -0.476 & -0.553 & -0.553 \\
    Materials Sci. & -0.228 & -0.289 & -0.315 \\
    Mathematics & -0.364 & -0.460 & -0.486 \\
    Medicine & -0.453 & -0.522 & -0.521 \\
    Physics & -0.383 & -0.460 & -0.485 \\
    Psychology & -0.409 & -0.513 & -0.486 \\
    \bottomrule
  \end{tabular}
    \caption{Spearman correlation ($\rho$) between citation count and HR$_2$ by field and model. All $p < 0.001$.}

  \label{tab:spearman_all_models}
\end{table}

\begin{figure}[htbp]
  \centering
  \includegraphics[width=\columnwidth]{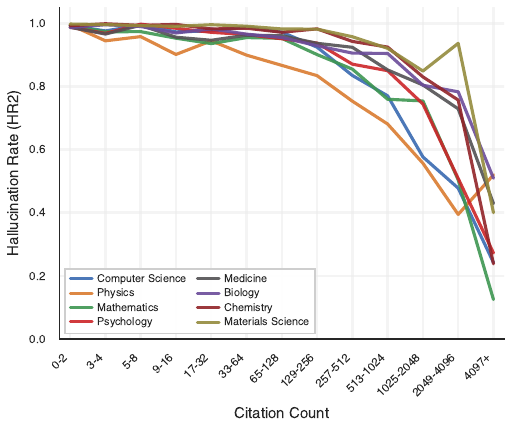}
  \captionsetup{font=small}
  \caption{Hallucination rate (HR$_2$) by citation bin across fields for GPT-4o.}
  \label{fig:field_hr_by_citations}
\end{figure}

Figure~\ref{fig:field_hr_by_citations} shows HR$_2$ vs. citation count
by field (GPT-4o). All fields trend downward from near-ceiling rates
at 0--2 citations, diverging at higher counts. Fluctuations at the highest citation counts reflect irreducible data sparsity (a small number of highly cited papers in the full 1.9M-paper dataset), rather than insufficient sampling.

Additional statistics based on all three definitions of hallucination rates across models, along with field-level performance breakdowns, are reported in Appendix~\ref{sec:appendix_results_hr_extra} and further corroborate our observations. Taken together, these results support our hypothesis that \emph{facts appearing more frequently in the training data are recalled more reliably by LLMs}. The consistency of this pattern across all eight fields and three models suggests that it reflects a fundamental property of how LLMs encode factual knowledge.

\section{Multiple-choice Recognition Analysis}
\label{sec:mc_recognition}

\begin{figure}[t]
\centering
\includegraphics[width=\columnwidth]{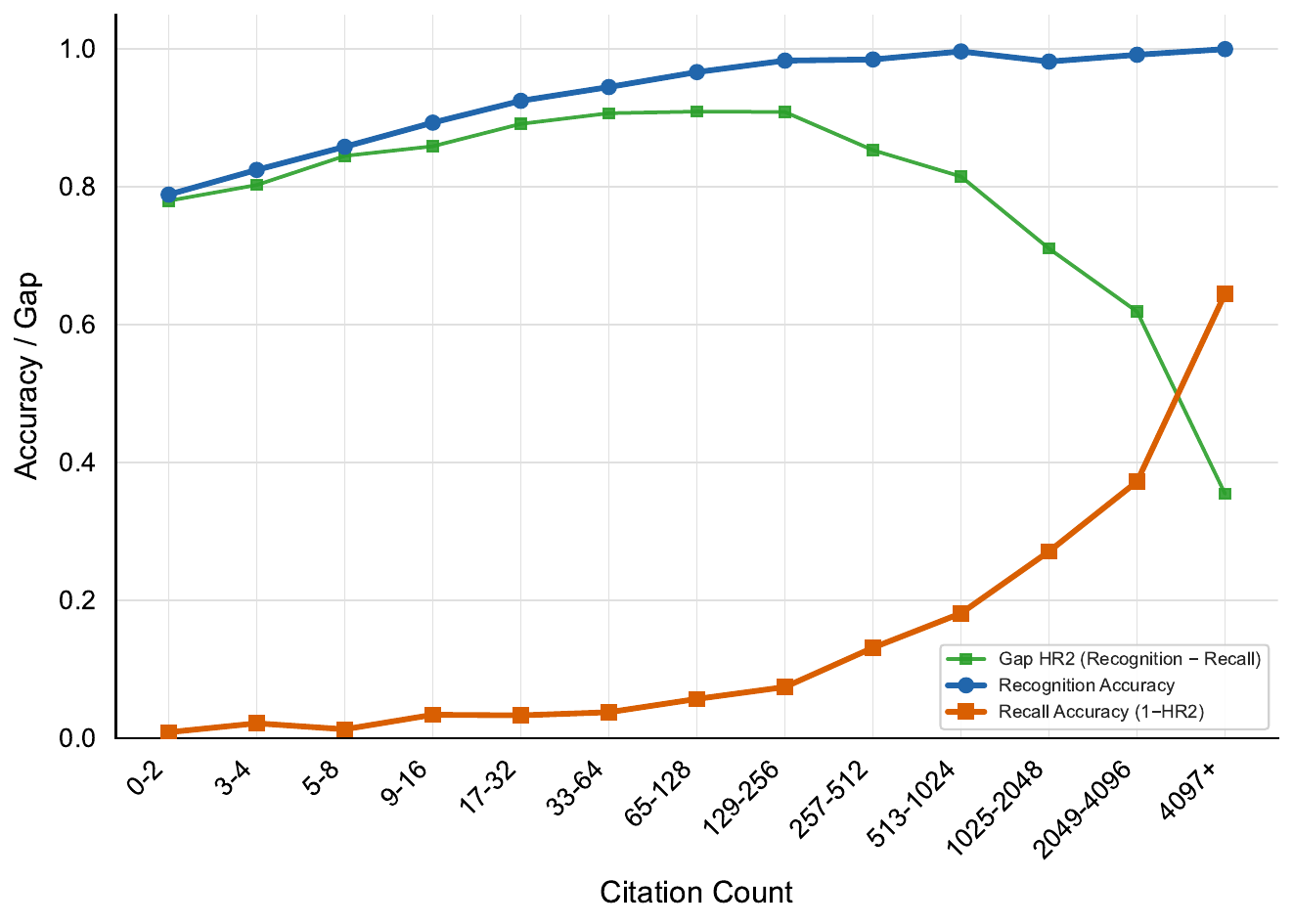}
\captionsetup{font=small}
\caption{Recall accuracy ($1 - \text{HR}_2$), chance-corrected MC recognition accuracy, and the gap between them across citation bins.}
\label{fig:mc_recognition_results}
\end{figure}

To test whether the hallucination failures documented in Section~\ref{sec:results} reflect genuine knowledge absence or a retrieval bottleneck, we conducted a multiple-choice (MC) recognition experiment using GPT-4o on the same 9{,}108 papers. Each paper was presented as a 4-option forced-choice task: the correct complete author list versus three same-field distractor teams of equal size. Figure~\ref{fig:mc_recognition_results} plots recall accuracy ($1 - \text{HR}_2$) and \emph{chance-corrected} MC accuracy (Eq.~\ref{eqn:chance_corrected}) across citation bins. Even for papers with 0--2 citations, where open-ended recall approaches zero, MC recognition achieves 79\% chance-corrected accuracy,
and this recognition advantage persists across the full citation spectrum.
This inverted-U trajectory is consistent with the qualitative structure of two-process (generate-recognize) accounts of recall \cite{anderson1972recognition}, in which free recall requires both retrieving a candidate from memory and recognizing it as correct, whereas a recognition task bypasses the retrieval stage entirely. Under this view, the recall–recognition gap should be small when the underlying memory representation is too weak to support recognition either, and should also shrink once the representation is strong enough for direct retrieval, leaving the largest gap at intermediate memory strength.
This indicates that the model's parametric knowledge of author teams is richer than free-form generation can surface: prevalence-driven hallucination primarily reflects a retrieval failure rather than complete knowledge absence.
To address metric commensurability, we also report results under a binary perfect-recall criterion ($H=U=0$; Appendix~\ref{sec:appendix_binary_recall}), since MC accuracy is binary whereas $1-\text{HR}_2$ allows partial credit. The recall--recognition gap is even larger under this stricter definition.
We also repeated the experiment with two more challenging distractor constructions: one in which each distractor shares $k-1$ of the $k$ true authors, and another in which a real collaborator of the true first author is substituted.
Recognition is essentially unchanged under the first and drops under the second, but stays far above chance in both, so the recall--recognition gap holds under harder controls at a more conservative magnitude (Appendix~\ref{sec:appendix_hard_distractors}).
Full experimental details are given in Appendix~\ref{sec:appendix_mc_recognition}.

\section{Mechanistic Evidence for Prevalence Encoding}
\label{sec:prevalence_encoding}

The behavioral results above show that prevalence is associated with both hallucination rate and recognition accuracy. We now ask whether this signal is also visible in LLMs' internal representations. Because the closed-source models used in Section~\ref{sec:results} do not expose internal activations, we conduct the hidden-state analyses on two open-source models, Qwen3-32B and Mistral-Small-3.2-24B, evaluated on the same 9{,}108 papers. These analyses therefore provide convergent evidence for the behavioral results rather than a direct account of the closed-source models themselves.
We find that the last prompt token before generation begins (P1) carries the strongest prevalence signal: a cross-validated ridge regression on P1 hidden states predicts $\log(\text{citations} + 1)$ with $R^2 = 0.615$ (Qwen3-32B) and 0.585 (Mistral), compared to an $R^2$ of 0.13 from a title-only baseline (TF-IDF + SBERT) and just 0.039 when using paper metadata features (e.g., field, year, author count, title length) alone.

\begin{figure}[h]
\centering
\includegraphics[width=\columnwidth]{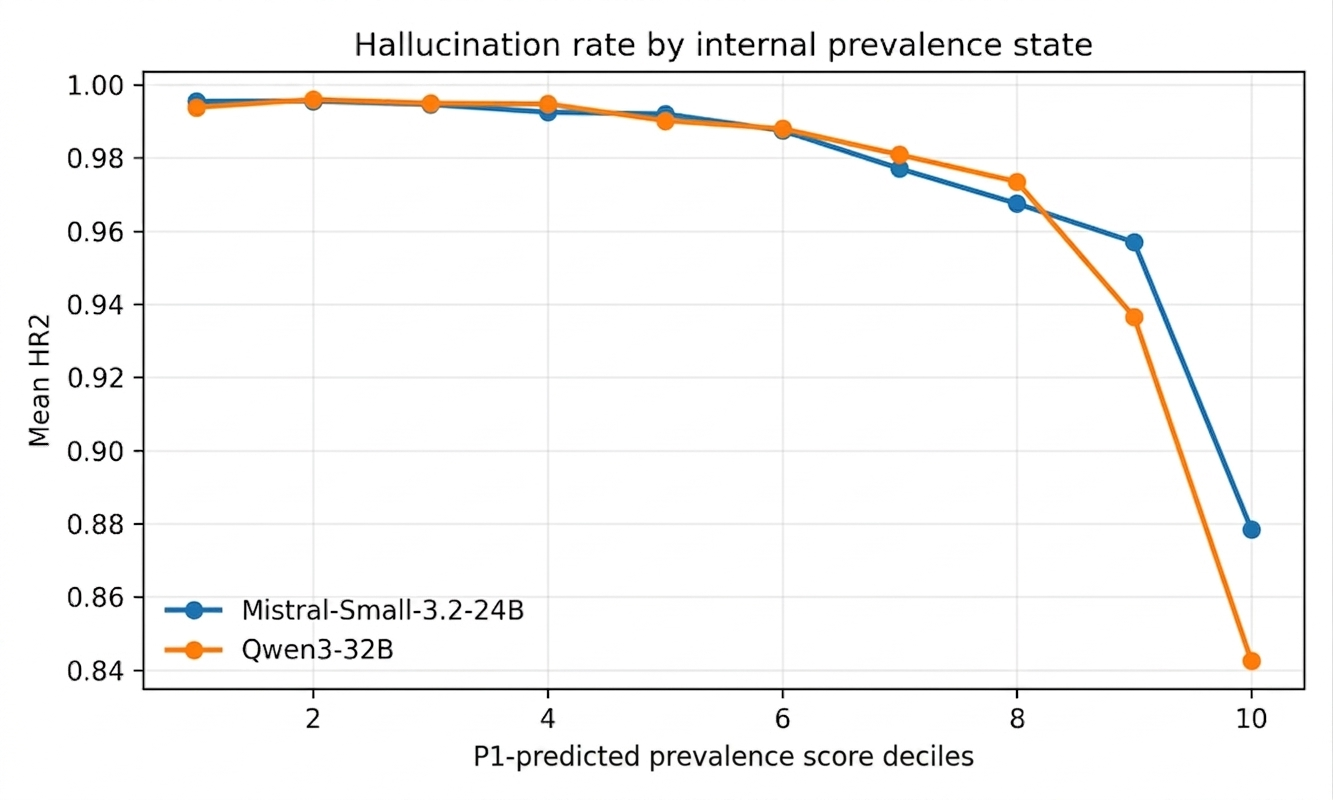}
\captionsetup{font=small}
\caption{Mean HR$_2$ by P1-decoded prevalence decile for Qwen3-32B and Mistral-Small-3.2-24B.}
\label{fig:internal_prevalence_deciles}
\end{figure}

As shown in
Figure~\ref{fig:internal_prevalence_deciles}, papers whose P1 representations encode higher prevalence have substantially lower hallucination rates:
mean $\text{HR}_2$ decreases from the lowest decoded-prevalence decile to the highest. A mediation-style decomposition using the fold-safe P1-decoded prevalence score estimates that this internal prevalence signal carries 67.3\% of the citation--hallucination association for Qwen3-32B and 51.9\% for Mistral-Small-3.2-24B. These results support the interpretation that pre-generation representations contain prevalence-related information linked to hallucination risk. An activation-steering intervention at the peak-decodability layer further shows that suppressing this direction increases hallucination on papers the model otherwise attributes correctly, beyond norm-matched random directions (Appendix~\ref{sec:appendix_steering}). Full
analyses are provided in Appendix~\ref{sec:appendix_prevalence_encoding}.

\section{Additional Findings}
\label{sec:findings}

Several supplementary analyses deepen our understanding of hallucination patterns while supporting the prevalence hypothesis.

\textbf{Author position.} Among responses in which at least one author is correct, recall degrades with position in the true author list: GPT-4o misses the first author in 41.7\% of cases and the tenth in 70.5\%. A parallel pattern holds by position in the LLM-generated author list, with GPT-4o's hallucination rate rising from 34.2\% at the first position to 75.5\% at the tenth; senior (last) authors show no recall advantage. Models can often name the primary contributor, but they rarely reconstruct the full team (Appendix~\ref{sec:author_position_hallucination}).

\textbf{Author count.} Hallucination rate correlates negatively with the number of true authors, which would suggest that larger teams are easier to recall. The mechanism is instead one of over-production: models pad short author lists, so single-author papers elicit 2.5 generated names of which 2.3 are fabricated, or 2.3 per true author, whereas papers with eleven or more authors elicit 10.0 names with 7.0 fabricated, or 0.4 per true author. Output length does track the true count ($\rho \approx 0.55$) and adjusts by field, but not steeply enough to hold the per-author error constant (Appendix~\ref{sec:author_count_hallucination}).

\textbf{Title length.} Longer titles are weakly associated with higher hallucination rates ($\rho = +0.035$ to $+0.050$, $p < 0.001$), but the association is carried by citation count: partial correlations that control for citations fall to $+0.007$ to $+0.019$ and none remains significant, so title length has no reliable effect of its own (Appendix~\ref{sec:title_length}).

\textbf{Multivariate controls.} A binomial-logit GLM that jointly controls for author count, title length, publication year, and field leaves the citation effect intact under both HR$_2$ and HR$_4$: each doubling of citations reduces the odds of hallucination by 34--36\%, while title length is not significant under either metric ($p > 0.37$). A mixed-effects model with a random intercept per field reaches the same conclusion, with an intraclass correlation of roughly 2\%, so unobserved field-level factors account for little residual variance (Appendix~\ref{sec:regression}).

\textbf{Prompt sensitivity.} On a stratified subsample of 1{,}004 papers, replacing the task-oriented prompt with an emotional or an authoritative variant leaves DeepSeek-R1's citation--hallucination correlation essentially unchanged (HR$_2$: $\rho = -0.450$ at baseline versus $-0.474$ for both variants). Paired bootstrap tests place both differences within noise, with 95\% confidence intervals that contain zero (Appendix~\ref{sec:appendix_prompt_sensitivity}).

\textbf{Ethnic composition of hallucinated names.} Fabricated names are close to representative in aggregate but not in variety. Weighted by how often a reader encounters them, their ethnic distribution closely tracks the academic population (Cram\'er's $V = 0.031$--$0.035$; Chinese names are 16.2\% of dataset occurrences and 17.3\% of hallucination instances). Counting each distinct fabricated name once, a small skew appears ($V = 0.09$--$0.10$): English/Western names are over-represented by 1.5--1.6$\times$, Korean by 1.4--1.8$\times$, and Japanese by 1.4--1.6$\times$, while Russian (0.60--0.67$\times$) and French (0.73--0.75$\times$) names are under-represented. The models thus draw on a narrower pool of distinct names for some groups than the literature itself offers. The pattern holds across three architecturally distinct models, pointing to shared properties of web-scale training data rather than a model-specific artifact (Appendix~\ref{sec:ethnicity}).

\section{Conclusion}
\label{sec:conclusion}

In this study, we systematically investigate the relationship between factual prevalence and LLM hallucination in academic authorship attribution, using a large-scale dataset curated across eight disciplines and thirteen citation bins. Across three state-of-the-art LLMs, we find a universal inverse relationship between citation count and hallucination rate, with consistently high error rates for low-citation papers across all models and fields.

This prevalence-driven hallucination is a fundamental property of current LLMs, not merely an artifact of any specific model or domain. Multiple-choice recognition experiments under strict distractor controls further show that these failures primarily reflect retrieval limitations rather than complete knowledge absence. Hidden-state probing adds convergent mechanistic evidence, suggesting that prevalence is encoded in pre-generation representations that align with our behavioral findings.

These results point toward retrieval-augmented and prevalence-aware strategies for improving LLM factual reliability.

\section*{Limitations}
\label{sec:limitations}

While our study provides robust evidence for the inverse relationship between factual prevalence and hallucination rates, several limitations warrant consideration:

\textbf{Proxy Imperfection \& Data Availability.} We use citation count as a proxy for training-data prevalence and validate this choice by showing that citation count is proportional to both Google search hits and the frequency of corresponding $n$-grams in two large web corpora. Ideally, prevalence would be measured directly from the LLMs' training data; however, the training corpora of industrial LLMs remain opaque to academic researchers.

\textbf{Temporal Misalignment \& Knowledge Lag.}
Our dataset includes papers published up to 2019 to accommodate model knowledge cutoffs, with citation counts obtained from Semantic Scholar as of September 2025. However, citation counts are inherently dynamic. As a result, a paper could have accrued a substantial portion of its citations after the LLM's training cutoff, leading to a mismatch between its citation-based prevalence score and its true frequency in the model's training data.

\textbf{Forced-Answer Protocol.} Our prompt requires a comma-separated author list and offers no abstention option, so a model that lacks the relevant knowledge can only produce names. The reported rates therefore measure error under forced answering and may overstate how often a deployed assistant, which can decline to answer, would volunteer fabricated authors. Two results limit this concern: HR$_4$ isolates fabricated names from omitted ones and follows the same citation trend (Appendix~\ref{sec:appendix_HR_definitions}), and the recognition results in Section~\ref{sec:mc_recognition} show that the failures are not a simple absence of knowledge. Measuring how much of the long-tail error converts to abstention under an explicit ``unknown'' option is valuable future work.

\textbf{Metadata Noise.} While we applied rigorous filtering to the Semantic Scholar dataset, academic metadata inherently contains noise (e.g., author name disambiguation errors, inconsistent middle initials). Although our hard-coded evaluation method mitigates this via fuzzy string matching, some ground-truth errors likely persist.

\textbf{Scope of Hidden-State Analysis.} Our hidden-state probing (Section~\ref{sec:prevalence_encoding}) is conducted on two open-source models (Qwen3-32B and Mistral-Small-3.2-24B), which differ from the closed-source models evaluated behaviorally (GPT-4o, DeepSeek-R1, Claude Sonnet 4.5). Whether the same prevalence-encoding mechanism operates in closed-source architectures, which may differ substantially in training data, scale, and alignment procedures, remains an open question. In addition, while the P1 prevalence mediator accounts for 51.9--67.3\% of the citation--hallucination association (51.9\% for Mistral-Small-3.2-24B and 67.3\% for Qwen3-32B), the remaining direct path indicates that additional mechanisms beyond those captured by our probes also drive hallucination patterns.

\textbf{Closed-Book Setting vs RAG.} Our evaluation is deliberately closed-book: API-based queries of the kind we issue do not trigger Retrieval-Augmented Generation (RAG), which isolates parametric knowledge and establishes the baseline that any retrieval-based mitigation must improve upon. Our results should therefore not be read as error rates for deployed assistants that can search the web or query citation databases, where authorship is often a direct retrieval target. Retrieval does not by itself remove the problem, since retrieval-augmented and search-based assistants still produce reference errors at non-trivial rates~\citep{rao2026reference}, and author lists and ordering can differ across versions of the same paper, so a retrieved record does not guarantee the canonical attribution. We do not evaluate retrieval-augmented systems here, and whether RAG can fully eliminate the long-tail hallucination penalty or merely shifts errors from parametric memory to context-window confusion remains an open and important question.

Building on these findings, we propose \textbf{three key avenues for future research}: 1) Moving beyond proxies to directly correlate hallucination rates with the frequency of specific entity tokens in training corpora (e.g., CommonCrawl). 2) Developing ``prevalence-aware'' systems that use metadata (like citation counts) to dynamically adjust temperature or trigger RAG for low-prevalence queries.
3) Explicitly testing whether RAG can resolve the retrieval bottleneck identified in our recognition experiment, or whether retrieval noise shifts errors from parametric memory to context-window confusion, and investigating whether architectural or training interventions can reduce prevalence encoding in model representations.

\section*{Ethical Considerations}

Our study uses publicly available bibliographic metadata from the Open Academic Graph and Semantic Scholar. These records contain author names, which are personal data. We use them only to check whether a model reproduces a paper's published author list, and we report all results in aggregate. We do not evaluate or rank individual researchers, and a hallucinated name produced by a model is not a claim about the person who bears that name.

The ethnicity distribution analysis in Appendix~\ref{sec:ethnicity} infers a coarse category from surnames alone. Surnames are a noisy proxy: they do not capture self-identification, they conflate national origin with ethnicity, and they misclassify names that fall between the classifier's categories. We therefore use the classifier only to compare aggregate distributions between the dataset and the hallucinated names, never to label an individual, and we treat the resulting effect sizes as descriptive.

\section*{Acknowledgments}

We thank the anonymous reviewers and the Area Chair of ACL Rolling Review for their careful and constructive feedback. Their suggestions made this work substantially more robust. We also acknowledge the use of generative AI tools for language polishing and for condensing the text.

\bibliography{custom}

\appendix
\setcounter{figure}{0}
\renewcommand{\thesection}{A\arabic{section}}
\setcounter{section}{0}
\renewcommand{\thetable}{A\arabic{table}}
\setcounter{table}{0}
\renewcommand{\thefigure}{A\arabic{figure}}
\setcounter{figure}{0}
\renewcommand{\theequation}{A\arabic{equation}}
\setcounter{equation}{0}

\section*{Appendix}

\section{Related work}
\label{sec:appendix_related_work}

Hallucination in LLMs is a well-known issue that has attracted much attention, because of its potential adverse impact in real-world applications. For example, hallucination in the context of medical literature was studied by \citet{Chelli2024}. The study found that using LLMs such as ChatGPT to conduct systematic reviews for medical conditions can generate misleading or “hallucinated” references, exceeding a 25\% rate. \citet{zheng2023doeschatgptfallshort} identifies four error types of ChatGPT and pinpoints factuality hallucination as the most critical error. This is the hallucination we are studying in this paper. A number of survey papers have been written~\citep{Huang_2025_hallucination_survey, zhang2025sirenssongaiocean} about LLM hallucination. In this section, we highlight works that we consider directly related to our work, but refer to the survey papers for additional topics, such as the categorization of different hallucinations.

There is a large body of work on detecting hallucinations. For instance, \citet{Kadavath2022} explores whether language models ``know what they know,'' finding that via a measure known as P(True), LLMs can self-evaluate confidence of facts with encouraging performance on a diverse array of tasks. SelfCheckGPT \citep{Manakul2023} detects hallucinations via cross-checking the consistency of multiple outputs.
Similarly, \citet{farquhar2024} used semantic entropy to capture uncertainty via entailment-based clustering, which is suited for free-form generation, though its effectiveness depends on clustering quality.
\citet{phillips2025geometricuncertaintydetectingcorrecting} proposed to use the volume of the convex hull of the LLM response embeddings as a measure of uncertainty.
In contrast to these, our study investigates the cause, specifically the role of factual correctness in authorship questions. Because of our specific setup, we match LLM responses with the ground truth using either LLM self-evaluation or direct algorithmic string matching, which are more straightforward and robust.

There has been much work on how LLMs capture factual knowledge. \citet{li2022} conduct a causal-inspired analysis of how pre-trained language models capture factual knowledge. They find that models often rely on shallow patterns (e.g., association or positional proximity) in data rather than deep semantic understanding. \citet{akyurek2022} studied methods to trace factual knowledge back to training data, showing that traditional retrieval methods, such as BM25, are better than gradient or embedding-based methods.

Causes of hallucination have also been studied in the literature. \citet{dziri2022} analyzed knowledge-grounded conversational models and attributed hallucinations to inconsistencies and biases in datasets, such as subjective information (e.g., personal opinions) and unsupported objective facts (invented or unverified facts). \citet{onoe-etal-2022-entity} demonstrated
that handling completely unseen entities remains
challenging for LLMs. \citet{mckenna2023sourceshallucinationlargelanguage} point out two pretraining biases that make LLMs hallucinate in natural language inference: attestation bias, where models wrongly say a statement is entailed about twice as often, and relative frequency bias, which raises errors 1.6–2× when the premise's words appear less often than the hypothesis's. These biases make models look good on standard data but perform almost randomly on tricky examples, showing they rely on shallow patterns instead of logical reasoning.

Several previous works have studied the memorization capabilities of LLMs. \citet{carlini2022quantifying} demonstrated that memorization improves with model size and with the number of times an example is duplicated in the training data, while \citet{razeghi2022_pretrainingTermFreqFewShot} showed that few-shot reasoning accuracy tracks how often the relevant terms appear in the pretraining corpus. \citet{Kandpal2022_LLMStruggleLongTail} found that a language model's performance on factoid QA datasets correlates with the number of pretraining documents relevant to the question. \citet{mallen2023_whenNotToTrustLMs} analyzed knowledge questions from Wikipedia and observed that LLMs struggle with factual knowledge for pages with fewer views. \citet{sun2024headtotail} introduced a benchmark to assess LLMs' ability to internalize head, torso, and tail facts, showing that even state-of-the-art models have significant limitations, particularly for torso and tail entities. Our work differs in that we focus specifically on the \emph{academic authorship attribution} task and establish a correlation between hallucination rates and citation counts. We further validate citation counts as a proxy for factual prevalence by examining their correlation with Google search hits and frequency in Infini-gram~\citep{Liu2024InfiniGram}. Additionally, we investigate domain variability and ethnicity bias in hallucinated authors, providing further insight into systematic patterns in LLM errors.

Finally, there are a number of datasets designed for hallucination detection. For example, \citet{Wei2024} proposed SimpleQA to benchmark factual accuracy in short-form QA. While their benchmark is useful for atomic facts, it does not address structured or multi-entity responses. Our work extends this line by curating a dataset of academic literature and examining how hallucination varies with topic familiarity in authorship-related questions, using citation count as a proxy for topic prevalence across papers in STEM fields.

\section{Dataset Details}

\label{sec:appendix_dataset}

\subsection{Details of Data Filtering}
\label{sec:appendix_filtering}

This section provides the complete multi-stage data filtering procedure used to curate the final dataset from an initial corpus of 9,999,863 papers.

The filtering steps were executed in the following order:
\begin{enumerate}[
   leftmargin=1em,
    itemsep=0.2em,
    topsep=0.2em,
    partopsep=0em,
    parsep=0em
]
    \item \textbf{Filter by Document Type:} We first narrowed our dataset by document type, retaining only entries classified as ``Journal'' or ``Conference'' papers. This step removed other publication types like Book, BookChapter, and Patent, filtering out 5,686,732 records (56.9\% of the initial dataset).

    \item \textbf{Filter by Author Count:} We filtered out any records that had no authors or more than 20 authors. This removed 14,874 records (0.3\%), with the vast majority of papers falling within the 1-20 author range.

    \item \textbf{Filter by Language:} We utilized the \texttt{fasttext} library to perform language identification. Only papers with titles identified as being written in English were kept. This was the most significant quality filter, removing 2,058,086 records (47.9\% of papers that passed the previous filters).

    \item \textbf{Filter by Author Name Patterns:} We applied a pattern validation filter to remove records with problematic author name formats, such as malformed entries or non-standard characters. This step removed 18,660 records (0.8\%).

    \item \textbf{Verify Title and Author Match:} As a final quality control step, we verified the match between the original OAG record and the data retrieved from Semantic Scholar. We only retained entries where the lowercased and whitespace-stripped titles matched exactly, and the first listed author's last name matched exactly. This verification step removed 300,302 records (13.5\%).
\end{enumerate}

This rigorous filtering process ensured a high degree of confidence that the enriched data accurately corresponded to the originally sampled papers, resulting in a final curated dataset of 1,921,209 high-quality records (19.2\% of the initial dataset).

\usetikzlibrary{shapes.geometric, arrows.meta, positioning, calc, shadows}

\tikzset{
    filterbox/.style={rectangle, rounded corners, minimum width=4.5cm, minimum height=1.3cm, text centered, text width=4.3cm, align=center, draw=blue!70, fill=blue!15, font=\normalsize, drop shadow},
    databox/.style={rectangle, rounded corners, minimum width=4.5cm, minimum height=0.9cm, text centered, text width=4.3cm, align=center, draw=black!70, fill=gray!10, font=\normalsize, drop shadow},
    arrow/.style={->,>=stealth, line width=1.2pt, color=black!70},
}

\begin{center}

\begin{tikzpicture}[node distance=0.8cm]

\node[databox, fill=green!15, draw=green!70] (initial) {
    \textbf{Initial Dataset}\\
    \textbf{9,999,863} records
};

\node[filterbox, below=of initial] (filter1) {
    \textbf{Document Type}\\
    {\small Journal \& Conference only}\\
    \textbf{4,313,131} retained\\
    {\small 5,686,732 removed (56.9\%)}
};

\node[filterbox, below=of filter1] (filter2) {
    \textbf{Author Count}\\
    {\small 1-20 authors}\\
    \textbf{4,298,257} retained\\
    {\small 14,874 removed (0.3\%)}
};

\node[filterbox, below=of filter2] (filter3) {
    \textbf{Language}\\
    {\small FastText model (English only)}\\
    \textbf{2,240,171} retained\\
    {\small 2,058,086 removed (47.9\%)}
};

\node[filterbox, below=of filter3] (filter4) {
    \textbf{Author Name Patterns}\\
    {\small Valid name formats}\\
    \textbf{2,221,511} retained\\
    {\small 18,660 removed (0.8\%)}
};

\node[filterbox, below=of filter4] (filter5) {
    \textbf{Title/Author Match}\\
    {\small Cross-source verification}\\
    \textbf{1,921,209} retained\\
    {\small 300,302 removed (13.5\%)}
};

\node[databox, below=of filter5, fill=yellow!20, draw=orange!70] (final) {
    \textbf{Final Dataset}\\
    \textbf{1,921,209} records\\
    {\small \textit{19.2\% of original}}
};

\draw[arrow] (initial) -- (filter1);
\draw[arrow] (filter1) -- (filter2);
\draw[arrow] (filter2) -- (filter3);
\draw[arrow] (filter3) -- (filter4);
\draw[arrow] (filter4) -- (filter5);
\draw[arrow] (filter5) -- (final);

\end{tikzpicture}
{\captionsetup{hypcap=false}%
\captionof{figure}{Multi-stage filtering pipeline for dataset curation. Each filter progressively refines data quality: (1) Document Type keeps only journal and conference papers; (2) Author Count restricts to 1-20 authors; (3) Language uses FastText to identify English papers; (4) Author Name Patterns removes problematic name formats; (5) Title/Author Match verifies cross-source consistency. The final dataset retains 19.2\% of the original records.}\label{fig:data_filtering}}

\end{center}

\subsection{Dataset Distribution Statistics}
\label{sec:appendix_stats}

To provide context on the characteristics of the underlying academic graph from which we sampled, this section presents distribution statistics for key metadata fields. The following tables (Tables \ref{tab:citations-dist-sampled}, \ref{tab:authors-dist-sampled}, and \ref{tab:fos-dist-sampled}) are based on a random sample of 100,000 papers from our initial data pool before the filtering steps described in Appendix~\ref{sec:appendix_filtering}.

\begin{table}[htbp]
\centering

\begin{tabular}{lrrr}
\toprule
\textbf{Range} & \textbf{Count} & \% & \textbf{Cumul}\% \\
\midrule
0 & 20,911 & 20.91 & 20.91 \\
1 & 7,835 & 7.83 & 28.75 \\
2-4 & 12,890 & 12.89 & 41.64 \\
5-9 & 12,492 & 12.49 & 54.13 \\
10-19 & 14,328 & 14.33 & 68.46 \\
20-49 & 17,184 & 17.18 & 85.64 \\
50-99 & 8,179 & 8.18 & 93.82 \\
100-199 & 4,018 & 4.02 & 97.84 \\
200-499 & 1,704 & 1.70 & 99.54 \\
500-999 & 329 & 0.33 & 99.87 \\
1000-1999 & 92 & 0.09 & 99.96 \\
2000-4999 & 31 & 0.03 & 99.99 \\
5000-9999 & 7 & 0.01 & 100.00 \\
10000+ & 0 & 0.00 & 100.00 \\
\bottomrule
\end{tabular}
\caption{Citation distribution across a random sample of 100,000 papers.}
\label{tab:citations-dist-sampled}
\end{table}

\begin{table}[htbp]
\centering

\begin{tabular}{lrrr}
\toprule
\textbf{Authors} & \textbf{Count} & \% & \textbf{Cumul}\% \\
\midrule
1 & 27,078 & 27.08 & 27.08 \\
2 & 20,215 & 20.22 & 47.30 \\
3 & 16,246 & 16.25 & 63.55 \\
4 & 11,962 & 11.96 & 75.51 \\
5 & 8,195 & 8.20 & 83.71 \\
6 & 5,661 & 5.66 & 89.37 \\
7 & 3,527 & 3.53 & 92.90 \\
8 & 2,413 & 2.41 & 95.31 \\
9 & 1,565 & 1.57 & 96.88 \\
10 & 1,105 & 1.10 & 97.98 \\
11 & 605 & 0.60 & 98.58 \\
12 & 477 & 0.48 & 99.06 \\
13 & 277 & 0.28 & 99.34 \\
14 & 228 & 0.23 & 99.57 \\
15 & 131 & 0.13 & 99.70 \\
16 & 90 & 0.09 & 99.79 \\
17 & 68 & 0.07 & 99.86 \\
18 & 69 & 0.07 & 99.93 \\
19 & 45 & 0.04 & 99.97 \\
20 & 43 & 0.04 & 100.01 \\
\bottomrule
\end{tabular}
\caption{Author count distribution across a random sample of 100,000 papers.}
\label{tab:authors-dist-sampled}
\end{table}

\begin{table}[htbp]
\centering

\begin{tabular}{lrrr}
\toprule
\textbf{Field} & \textbf{Count} & \% & \textbf{Cumul}\% \\
\midrule
Medicine & 26,622 & 26.62 & 26.62 \\
Chemistry & 12,297 & 12.30 & 38.92 \\
Biology & 11,623 & 11.62 & 50.54 \\
Materials Sci. & 7,425 & 7.42 & 57.96 \\
Comp. Sci. & 7,013 & 7.01 & 64.97 \\
Physics & 6,076 & 6.08 & 71.05 \\
Psychology & 5,071 & 5.07 & 76.12 \\
Mathematics & 4,298 & 4.30 & 80.42 \\
Engineering & 3,739 & 3.74 & 84.16 \\
Sociology & 2,658 & 2.66 & 86.82 \\
Economics & 2,258 & 2.26 & 89.08 \\
Geology & 2,013 & 2.01 & 91.09 \\
Unknown & 1,661 & 1.66 & 92.75 \\
Environ. Sci. & 1,582 & 1.58 & 94.33 \\
Business & 1,356 & 1.36 & 95.69 \\
Political Sci. & 1,235 & 1.23 & 96.92 \\
History & 1,170 & 1.17 & 98.09 \\
Geography & 854 & 0.85 & 98.94 \\
Philosophy & 566 & 0.57 & 99.51 \\
Art & 483 & 0.48 & 99.99 \\
\bottomrule
\end{tabular}
\caption{Distribution of field of study across a random sample of 100,000 papers.}
\label{tab:fos-dist-sampled}
\end{table}

The following tables (Tables \ref{tab:fos-dist}, \ref{tab:authors-dist}, \ref{tab:temporal-dist}, \ref{tab:authors-by-field-dist}, and \ref{tab:citations-by-field-dist}) present distribution statistics for the complete dataset of 1.92 million papers after applying the filtering steps detailed in Appendix~\ref{sec:appendix_filtering}. The distributions remain largely consistent with the 100k sample.

\begin{table}[htbp]
    \centering
    \normalsize

    \begin{tabular}{@{}lrrr@{}}
    \toprule
    \textbf{Field} & \textbf{Count} & \% & \textbf{Cumul}\% \\
    \midrule
        Medicine & 518,465 & 26.99 & 26.99 \\
        Biology & 211,479 & 11.01 & 38.00 \\
        Chemistry & 199,824 & 10.40 & 48.40 \\
        Comp. Sci. & 145,407 & 7.57 & 55.97 \\
        Materials Sci. & 119,242 & 6.21 & 62.18 \\
        Psychology & 103,942 & 5.41 & 67.59 \\
        Physics & 101,313 & 5.27 & 72.86 \\
        Engineering & 92,384 & 4.81 & 77.67 \\
        Mathematics & 78,625 & 4.09 & 81.76 \\
        Sociology & 61,537 & 3.20 & 84.96 \\
        Economics & 49,205 & 2.56 & 87.52 \\
        Business & 38,482 & 2.00 & 89.52 \\
        Geology & 36,248 & 1.89 & 91.41 \\
        Environ. Sci. & 33,531 & 1.75 & 93.16 \\
        Political Sci. & 32,385 & 1.69 & 94.85 \\
        History & 24,473 & 1.27 & 96.12 \\
        Geography & 19,255 & 1.00 & 97.12 \\
        Philosophy & 12,407 & 0.65 & 97.77 \\
        Art & 11,442 & 0.60 & 98.37 \\
        Unknown & 31,563 & 1.64 & 100.00 \\
    \bottomrule
    \end{tabular}
    \caption{Distribution of fields of study.}
    \label{tab:fos-dist}
\end{table}

\begin{table}[htbp]
  \centering
  \normalsize

  \begin{tabular}{@{}lrrr@{}}
    \toprule
    \textbf{No. of Authors} & \textbf{Count} & \% & \textbf{Cum.}\% \\
    \midrule
    1 & 604,251 & 31.5 & 31.5 \\
    2 & 361,654 & 18.8 & 50.3 \\
    3 & 294,182 & 15.3 & 65.6 \\
    4 & 214,592 & 11.2 & 76.8 \\
    5 & 147,576 & 7.7 & 84.4 \\
    6 & 101,932 & 5.3 & 89.7 \\
    7 & 65,272 & 3.4 & 93.1 \\
    8 & 43,571 & 2.3 & 95.4 \\
    9 & 27,908 & 1.5 & 96.9 \\
    10 & 20,607 & 1.1 & 97.9 \\
    11 & 12,093 & 0.6 & 98.6 \\
    12 & 8,609 & 0.4 & 99.0 \\
    13 & 5,593 & 0.3 & 99.3 \\
    14 & 3,975 & 0.2 & 99.5 \\
    15 & 2,807 & 0.1 & 99.7 \\
    16 & 2,030 & 0.1 & 99.8 \\
    17 & 1,545 & 0.1 & 99.9 \\
    18 & 1,153 & 0.1 & 100 \\
    19 & 980 & 0.1 & 100 \\
    20 & 879 & 0.0 & 100 \\
    \bottomrule
  \end{tabular}
    \caption{Distribution of author count.}
  \label{tab:authors-dist}
\end{table}

\begin{table}[htbp]
\centering

\begin{tabular}{@{}lrrr@{}}
\toprule
\textbf{Decade} & \textbf{Count} & \% & \textbf{Cumul}\% \\
\midrule
1800-1809 & 39 & 0.00 & 0.00 \\
1810-1819 & 49 & 0.00 & 0.00 \\
1820-1829 & 58 & 0.00 & 0.00 \\
1830-1839 & 124 & 0.01 & 0.01 \\
1840-1849 & 286 & 0.01 & 0.02 \\
1850-1859 & 556 & 0.03 & 0.05 \\
1860-1869 & 556 & 0.03 & 0.08 \\
1870-1879 & 1,083 & 0.06 & 0.14 \\
1880-1889 & 1,422 & 0.07 & 0.21 \\
1890-1899 & 2,022 & 0.11 & 0.32 \\
1900-1909 & 2,482 & 0.13 & 0.45 \\
1910-1919 & 3,203 & 0.17 & 0.62 \\
1920-1929 & 5,216 & 0.27 & 0.89 \\
1930-1939 & 9,165 & 0.48 & 1.37 \\
1940-1949 & 11,598 & 0.60 & 1.97 \\
1950-1959 & 23,633 & 1.23 & 3.20 \\
1960-1969 & 45,318 & 2.36 & 5.56 \\
1970-1979 & 93,357 & 4.86 & 10.42 \\
1980-1989 & 154,125 & 8.02 & 18.44 \\
1990-1999 & 256,674 & 13.36 & 31.80 \\
2000-2009 & 514,006 & 26.76 & 58.56 \\
2010-2019 & 796,237 & 41.44 & 100 \\
\bottomrule
\end{tabular}
\caption{Temporal distribution of publications (by decade).}
\label{tab:temporal-dist}
\end{table}

\begin{table}[htbp]
\centering

\begin{tabular}{@{}lrr@{}}
\toprule
\textbf{Field} & \textbf{Papers} & \textbf{Mean Authors} \\
\midrule
Medicine & 518,465 & 4.1 \\
Biology & 211,479 & 4.1 \\
Materials Sci. & 119,242 & 3.8 \\
Chemistry & 199,824 & 3.6 \\
Environ. Sci. & 33,531 & 3.2 \\
Unknown & 31,563 & 3.2 \\
Physics & 101,313 & 3.2 \\
Geology & 36,248 & 2.9 \\
Comp. Sci. & 145,407 & 2.7 \\
Geography & 19,255 & 2.4 \\
Psychology & 103,942 & 2.4 \\
Engineering & 92,384 & 2.3 \\
Mathematics & 78,625 & 2.1 \\
Business & 38,482 & 1.8 \\
Economics & 49,205 & 1.7 \\
Sociology & 61,537 & 1.4 \\
Political Sci. & 32,385 & 1.4 \\
Art & 11,442 & 1.2 \\
History & 24,473 & 1.2 \\
Philosophy & 12,407 & 1.1 \\
\bottomrule
\end{tabular}
\caption{Mean author count by field of study.}
\label{tab:authors-by-field-dist}
\end{table}

\begin{table}[htbp]
\centering

\begin{tabular}{@{}lrr@{}}
\toprule
\textbf{Field} & \textbf{Papers} & \textbf{Mean Citations} \\
\midrule
Biology & 211,479 & 45.0 \\
Psychology & 103,942 & 39.2 \\
Economics & 49,205 & 34.5 \\
Environ. Sci. & 33,531 & 33.0 \\
Medicine & 518,465 & 32.8 \\
Geology & 36,248 & 32.7 \\
Chemistry & 199,824 & 31.3 \\
Comp. Sci. & 145,407 & 26.9 \\
Materials Sci. & 119,242 & 26.0 \\
Mathematics & 78,625 & 25.4 \\
Physics & 101,313 & 24.9 \\
Business & 38,482 & 24.3 \\
Geography & 19,255 & 23.4 \\
Sociology & 61,537 & 18.4 \\
Engineering & 92,384 & 13.6 \\
Political Sci. & 32,385 & 12.0 \\
History & 24,473 & 5.7 \\
Philosophy & 12,407 & 5.6 \\
Unknown & 31,563 & 4.6 \\
Art & 11,442 & 3.6 \\
\bottomrule
\end{tabular}
\caption{Mean citation count by field of study.}
\label{tab:citations-by-field-dist}
\end{table}

\section{More Results on Citation As a Proxy}

\subsection{Google Search Hits Analysis}
\label{sec:appendix_web_presence}

We initially used the dataset of 6,048 papers for the Google Search query number-of-hits crawling process. However, due to various sources of instability during crawling, such as API timeouts, dynamic bot-detection triggers, inconsistent rendering of search result pages, and transient network latency issues, only 5,179 valid results were obtained. Most citation buckets retained more than 400 papers (except for buckets that originally contained fewer papers). After computing the relevant statistics for each bucket, we obtained the results shown in Table \ref{tab:google_hits_full}.

\begin{table}[h!]
\setlength{\tabcolsep}{1pt}
\centering

\resizebox{0.48\textwidth}{!}{%
\begin{tabular}{lcccccccc}
\toprule
\textbf{Citation} & \textbf{Count} & \textbf{Mean} & \textbf{CV} & \textbf{Min} & \textbf{Q25} & \textbf{Q50} & \textbf{Q75} & \textbf{Max}  \\
\midrule

0--2        & 417 & 197239.34 & 13.00 & 0 & 1 & 2 & 9 & 49300000\\
3--4        & 429 & 9843.02 & 12.77 & 0 & 1 & 3 & 8 & 2160000\\
5--8        & 430 & 95981.06 & 15.69 & 0 & 1 & 3 & 8 & 29400000\\
9--16       & 429 & 998.81 & 7.52 & 0 & 1 & 2 & 8 & 91600\\
17--32      & 413 & 395.15 & 9.94 & 0 & 1 & 3 & 8 & 65900\\
33--64      & 429 & 476.04 & 6.97 & 0 & 1 & 3 & 9 & 36600\\
65--128     & 435 & 414.48 & 9.33 & 0 & 1 & 3 & 9 & 67600\\
129--256    & 437 & 214.35 & 9.14 & 0 & 2 & 4 & 10 & 34200\\
257--512    & 434 & 191.73 & 9.42 & 0 & 2 & 5 & 56.75 & 36400\\
513--1024   & 416 & 152.13 & 3.25 & 0 & 3 & 6.5 & 111.50 & 6170\\
1025--2048  & 431 & 287.33 & 3.76 & 0 & 4 & 10 & 206.50 & 18200\\
2049--4096  & 362 & 10780.70 & 18.28 & 0 & 5 & 66 & 338.25 & 3750000\\
4097+       & 117 & 1415.34 & 2.60 & 0 & 8 & 182 & 1390 & 25600\\
\bottomrule
\end{tabular}
}%
\caption{Google search hits vs citation buckets (N = 5,179).}
\label{tab:google_hits_full}
\end{table}

The table shows that the mean number of hits fluctuates dramatically across buckets, particularly for the smallest citation buckets and the largest ones, making it difficult to discern any clear trend. Buckets 0–2, 3–4, 5–8, and 2049–4096 have extremely large coefficient of variation (CV) values, mostly above 10, implying severe fluctuation in the data. Therefore, it is evident that the dataset contains many outliers, especially in the smaller buckets. For example, in the 0–2 bucket, one paper with zero citations has a number of hits as high as 49,300,000.

We suspect that these cases are likely related to the length of the title+authors search query string. Some queries are very short, which makes them more likely to appear in unrelated webpages, thus inflating the number of irrelevant search results counted as hits. For instance, the paper with 49,300,000 hits has the query \texttt{"Is this tragic or comic" "H. White"}, which is short and likely to appear frequently across the web. To verify this, we computed the string length of each title+authors search query and calculated the average query length for each bucket, as shown in Table \ref{tab:query_length_full}. We found that the smallest bucket (0–2) indeed has the smallest average string length, approximately 28\% shorter than the largest bucket based on our dataset. The smallest three buckets and the largest two buckets all have relatively short average query lengths, suggesting that query length does influence the number of hits retrieved.

\begin{table}[h!]
\setlength{\tabcolsep}{1pt}
\centering

\begin{tabular}{lccc}
\toprule
\textbf{Citation} & \textbf{Count} &
\textbf{Avg Length} & \textbf{Std Length} \\
\midrule
0--2 & 417 & 105.80 & 50.42 \\
3--4 & 429 & 121.23 & 47.58 \\
5--8 & 430 & 122.20 & 52.51 \\
9--16 & 429 & 134.32 & 51.35 \\
17--32 & 413 & 139.16 & 54.42 \\
33--64 & 429 & 146.55 & 56.97 \\
65--128 & 435 & 143.89 & 53.98 \\
129--256 & 437 & 141.53 & 55.09 \\
257--512 & 434 & 137.92 & 59.36 \\
513--1024 & 416 & 128.15 & 57.53 \\
1025--2048 & 431 & 124.73 & 60.64 \\
2049--4096 & 362 & 115.58 & 57.77 \\
4097+ & 117 & 113.46 & 55.51 \\
\midrule
\textbf{Overall} & \textbf{5179} & \textbf{129.97} & \textbf{56.16} \\
\bottomrule
\end{tabular}
\caption{Average and standard deviation of search-query string length (title + authors) across citation buckets (N = 5,179).}
\label{tab:query_length_full}
\end{table}

Taken together, these outliers substantially affect the overall analysis and appear across multiple buckets. Therefore, to remove these outliers while preserving as much data as possible, we removed the extremely large hit values within each bucket, so as to reduce the CV of hits as much as possible. After several trials, we settled on removing hit values above the 80\% quantile in each bucket and used the resulting dataset for further analysis. The results are given in Table~\ref{tab:google_hits_filtered_80}, which show a clear correlation between the mean and median Google Search hits, and the citation count.

\begin{table}[h!]
\setlength{\tabcolsep}{1pt}
\centering

\resizebox{0.48\textwidth}{!}{%
\begin{tabular}{lcccccccc}
\toprule
\textbf{Citation} & \textbf{Count} & \textbf{Mean} & \textbf{CV} &
\textbf{Min} & \textbf{Q25} & \textbf{Q50} & \textbf{Q75} & \textbf{Max} \\
\midrule
0--2        & 333 & 3.22 & 2.05 & 0 & 0 & 1 & 4 & 58 \\
3--4        & 354 & 2.96 & 1.02 & 0 & 1 & 2 & 5 & 10 \\
5--8        & 348 & 2.87 & 1.03 & 0 & 0 & 2 & 5 & 10 \\
9--16       & 358 & 2.73 & 1.06 & 0 & 1 & 2 & 4 & 10 \\
17--32      & 341 & 2.94 & 0.99 & 0 & 1 & 2 & 5 & 10 \\
33--64      & 343 & 3.27 & 1.29 & 0 & 1 & 2 & 5 & 41 \\
65--128     & 352 & 3.35 & 0.88 & 0 & 1 & 2 & 5 & 10 \\
129--256    & 349 & 4.33 & 1.46 & 0 & 1 & 3 & 5 & 48 \\
257--512    & 347 & 10.33 & 1.86 & 0 & 2 & 4 & 7 & 91 \\
513--1024   & 333 & 22.57 & 1.79 & 0 & 2 & 5 & 10 & 170 \\
1025--2048  & 345 & 46.47 & 1.64 & 0 & 3 & 6 & 63 & 306 \\
2049--4096  & 289 & 82.90 & 1.41 & 0 & 4 & 8 & 130 & 449 \\
4097+       & 93 & 282.30 & 1.52 & 0 & 5 & 84 & 344 & 1630 \\
\bottomrule
\end{tabular}
}%
\caption{Google search hits vs citation buckets (N = 4,185).}
\label{tab:google_hits_filtered_80}
\end{table}

\subsection{LLM Corpus Occurrence Analysis}
\label{sec:appendix_corpus_occurrence}

Table~\ref{tab:infini-gram-olmo2-capitalize} and Table~\ref{tab:infini-gram-redpajama-capitalize} present detailed results of the Infini-gram $n$-gram count analysis using the OLMo2-32B and RedPajama corpora on the full dataset of 6,048 papers. Although the coefficients of variation (CV) are relatively high across citation buckets, both tables exhibit clear positive trends as citation counts increase.

To assess the sensitivity of these trends to sampling choice, we performed an additional independent random sampling of approximately 500 papers per citation bin and repeated the Infini-gram query procedure on the OLMo2-32B corpus. The resampled dataset produces a highly similar monotonic growth pattern in mean $n$-gram counts across citation buckets (Table~\ref{tab:infini-gram-olmo2-capitalize-2}). While exact bucket rankings vary (Spearman = 0.39), the overall log–log relationship and slope remain consistent, indicating robust scaling behavior.

We further explore an alternative query formulation in which the paper title is converted to Camel Case (with each word capitalized) and combined with the first author name using the logical operator \texttt{AND} (e.g., \texttt{Attention Is All You Need AND Ashish Vaswani}). Applying this query format to the OLMo2-32B and RedPajama corpora yields the results shown in Table~\ref{tab:infini-gram-olmo2-camel} and Table~\ref{tab:infini-gram-redpajama-camel}. We observe that the mean $n$-gram counts for all citation buckets are dramatically lower than those obtained using the capitalized title + first author query format (e.g., in the OLMo2-32B corpus, 661.23 vs.\ 18.08 for the 4097+ citation bucket). Moreover, for citation buckets below 1025--2048, the mean $n$-gram counts are nearly zero across most buckets, indicating that queries based on Camel Case titles occur extremely rarely and do not yield sufficient counts for reliable analysis. Correspondingly, CV values are exceptionally high, often exceeding 20, and in some cases are undefined due to zero variance.

Taken together, these results demonstrate that variations in query formulation can lead to highly unstable and incomplete $n$-gram counts, suggesting that using Infini-gram corpus counts as a prevalence proxy is unreliable and impractical.

\begin{table}[htb!]
\setlength{\tabcolsep}{1pt}
\centering

\resizebox{0.48 \textwidth}{!}{%
\begin{tabular}{lcccccccc}
\toprule
\textbf{Citation} &
\textbf{Count} &
\textbf{Mean} &
\textbf{CV} &
\textbf{Min} &
\textbf{Q25} &
\textbf{Q50} &
\textbf{Q75} &
\textbf{Max} \\
\midrule
0--2 & 499 & 0.02 & 18.49 & 0 & 0 & 0 & 0 & 9 \\
3--4 & 500 & 0.07 & 10.97 & 0 & 0 & 0 & 0 & 12 \\
5--8 & 500 & 0.21 & 15.86 & 0 & 0 & 0 & 0 & 75 \\
9--16 & 500 & 0.15 & 8.07 & 0 & 0 & 0 & 0 & 16 \\
17--32 & 500 & 0.50 & 8.38 & 0 & 0 & 0 & 0 & 80 \\
33--64 & 500 & 2.02 & 7.22 & 0 & 0 & 0 & 0 & 204 \\
65--128 & 500 & 3.25 & 11.96 & 0 & 0 & 0 & 0 & 819 \\
129--256 & 500 & 11.05 & 13.58 & 0 & 0 & 0 & 0 & 3321 \\
257--512 & 500 & 33.17 & 15.89 & 0 & 0 & 0 & 0 & 11763 \\
513--1024 & 500 & 34.41 & 4.44 & 0 & 0 & 0 & 7 & 1910 \\
1025--2048 & 500 & 148.49 & 7.45 & 0 & 0 & 0 & 11 & 16133 \\
2049--4096 & 415 & 124.47 & 5.53 & 0 & 0 & 0 & 23.50 & 12681 \\
4097+ & 134 & 661.23 & 5.49 & 0 & 0 & 4.5 & 208.75 & 37453 \\
\bottomrule
\end{tabular}
}%
\caption{Infini-gram $n$-gram count statistics across citation buckets using OLMo2-32B corpus (capitalized title + first author).}
\label{tab:infini-gram-olmo2-capitalize}
\end{table}

\begin{table}[htb!]
\setlength{\tabcolsep}{1pt}
\centering

\resizebox{0.48 \textwidth}{!}{%
\begin{tabular}{lccccccccc}
\toprule
\textbf{Citation} &
\textbf{Count} &
\textbf{Mean} &
\textbf{CV} &
\textbf{Std} &
\textbf{Min} &
\textbf{Q25} &
\textbf{Q50} &
\textbf{Q75} &
\textbf{Max} \\
\midrule
0--2 & 499 & 0.02 & 17.25 & 0.41 & 0 & 0 & 0 & 0 & 9 \\
3--4 & 500 & 0.05 & 7.89 & 0.39 & 0 & 0 & 0 & 0 & 4 \\
5--8 & 500 & 0.05 & 7.27 & 0.35 & 0 & 0 & 0 & 0 & 4 \\
9--16 & 500 & 0.07 & 8.22 & 0.54 & 0 & 0 & 0 & 0 & 8 \\
17--32 & 500 & 0.08 & 6.77 & 0.55 & 0 & 0 & 0 & 0 & 7 \\
33--64 & 500 & 0.27 & 7.40 & 2.03 & 0 & 0 & 0 & 0 & 32 \\
65--128 & 500 & 0.23 & 5.79 & 1.33 & 0 & 0 & 0 & 0 & 21 \\
129--256 & 500 & 0.78 & 6.48 & 5.08 & 0 & 0 & 0 & 0 & 99 \\
257--512 & 500 & 2.17 & 6.03 & 13.07 & 0 & 0 & 0 & 0 & 194 \\
513--1024 & 500 & 4.98 & 7.18 & 35.77 & 0 & 0 & 0 & 2 & 737 \\
1025--2048 & 500 & 8.01 & 5.58 & 44.71 & 0 & 0 & 0 & 4 & 905 \\
2049--4096 & 415 & 50.53 & 15.28 & 772.03 & 0 & 0 & 0 & 8 & 15727 \\
4097+ & 134 & 84.95 & 4.65 & 394.65 & 0 & 0 & 0 & 30.25 & 4230 \\
\bottomrule
\end{tabular}
}%
\caption{Infini-gram $n$-gram count statistics across citation buckets using RedPajama corpus (capitalized title + first author).}
\label{tab:infini-gram-redpajama-capitalize}
\end{table}

\begin{table}[htb!]
\setlength{\tabcolsep}{1pt}
\centering

\resizebox{0.48 \textwidth}{!}{%
\begin{tabular}{lcccccccc}
\toprule
\textbf{Citation} &
\textbf{Count} &
\textbf{Mean} &
\textbf{CV} &
\textbf{Min} &
\textbf{Q25} &
\textbf{Q50} &
\textbf{Q75} &
\textbf{Max} \\
\midrule
0--2 & 498 & 0.176 & 6.504 & 0.1 & 0.1 & 0.1 & 0.1 & 20.0 \\
3--4 & 500 & 0.139 & 5.756 & 0.1 & 0.1 & 0.1 & 0.1 & 18.0 \\
5--8 & 500 & 0.170 & 4.293 & 0.1 & 0.1 & 0.1 & 0.1 & 12.0 \\
9--16 & 500 & 0.308 & 9.278 & 0.1 & 0.1 & 0.1 & 0.1 & 61.0 \\
17--32 & 500 & 0.473 & 6.274 & 0.1 & 0.1 & 0.1 & 0.1 & 39.0 \\
33--64 & 500 & 1.934 & 7.846 & 0.1 & 0.1 & 0.1 & 0.1 & 284.0 \\
65--128 & 500 & 4.469 & 9.879 & 0.1 & 0.1 & 0.1 & 0.1 & 900.0 \\
129--256 & 500 & 33.774 & 15.926 & 0.1 & 0.1 & 0.1 & 0.1 & 11977.0 \\
257--512 & 500 & 54.195 & 13.849 & 0.1 & 0.1 & 0.1 & 0.325 & 16664.0 \\
513--1024 & 500 & 58.764 & 7.671 & 0.1 & 0.1 & 0.1 & 5.250 & 8301.0 \\
1025--2048 & 500 & 60.928 & 4.004 & 0.1 & 0.1 & 0.1 & 6.500 & 3208.0 \\
2049--4096 & 415 & 131.447 & 5.333 & 0.1 & 0.1 & 0.1 & 27.000 & 12765.0 \\
4097+ & 134 & 679.613 & 5.378 & 0.1 & 0.1 & 4.500 & 208.750 & 37645.0 \\
\bottomrule
\end{tabular}
}%
\caption{Infini-gram $n$-gram count statistics across citation buckets using OLMo2-32B corpus (capitalized title + first author), based on an independent resampling of approximately 500 papers per citation bin.}
\label{tab:infini-gram-olmo2-capitalize-2}
\end{table}

\begin{table}[htb!]
\setlength{\tabcolsep}{1pt}
\centering

\resizebox{0.48 \textwidth}{!}{%
\begin{tabular}{lccccccccc}
\toprule
\textbf{Citation} &
\textbf{Count} &
\textbf{Mean} &
\textbf{Std} &
\textbf{Min} &
\textbf{Q25} &
\textbf{Q50} &
\textbf{Q75} &
\textbf{Max} &
\textbf{CV} \\
\midrule
0--2     & 499 & 0      & 0       & 0 & 0 & 0 & 0 & 0    &      \\
3--4     & 500 & 0      & 0       & 0 & 0 & 0 & 0 & 0    &      \\
5--8     & 500 & 0   & 0.09    & 0 & 0 & 0 & 0 & 2    & 22.36 \\
9--16    & 500 & 0      & 0       & 0 & 0 & 0 & 0 & 0    &      \\
17--32   & 500 & 0   & 0.09    & 0 & 0 & 0 & 0 & 2    & 22.36 \\
33--64   & 500 & 0      & 0       & 0 & 0 & 0 & 0 & 0    &      \\
65--128  & 500 & 0      & 0       & 0 & 0 & 0 & 0 & 0    &      \\
129--256 & 500 & 0.05   & 1.16    & 0 & 0 & 0 & 0 & 26   & 21.55 \\
257--512 & 500 & 0.01   & 0.27    & 0 & 0 & 0 & 0 & 6    & 22.36 \\
513--1024 & 500 & 0.02  & 0.54    & 0 & 0 & 0 & 0 & 12   & 22.36 \\
1025--2048 & 500 & 0.33 & 4.82    & 0 & 0 & 0 & 0 & 97   & 14.69 \\
2049--4096 & 415 & 2.10 & 40.51   & 0 & 0 & 0 & 0 & 825  & 19.32 \\
4097+      & 134 & 18.08 & 173.66 & 0 & 0 & 0 & 0 & 2000 & 9.60  \\
\bottomrule
\end{tabular}
}%
\caption{Infini-gram $n$-gram count statistics across citation buckets using OLMo2-32B corpus (camelCase title + first author).}
\label{tab:infini-gram-olmo2-camel}
\end{table}

\begin{table}[htb!]
\setlength{\tabcolsep}{1pt}
\centering

\resizebox{0.48 \textwidth}{!}{%
\begin{tabular}{lccccccccc}
\toprule
\textbf{Citation Bucket} &
\textbf{Count} &
\textbf{Mean} &
\textbf{Std} &
\textbf{Min} &
\textbf{Q25} &
\textbf{Q50} &
\textbf{Q75} &
\textbf{Max} &
\textbf{CV} \\
\midrule
0--2      & 499 & 0    & 0     & 0 & 0 & 0 & 0 & 0    &      \\
3--4      & 500 & 0    & 0     & 0 & 0 & 0 & 0 & 0    &      \\
5--8      & 500 & 0    & 0     & 0 & 0 & 0 & 0 & 0    &      \\
9--16     & 500 & 0    & 0     & 0 & 0 & 0 & 0 & 0    &      \\
17--32    & 500 & 0    & 0     & 0 & 0 & 0 & 0 & 0    &      \\
33--64    & 500 & 0    & 0.04  & 0 & 0 & 0 & 0 & 1    & 22.36 \\
65--128   & 500 & 0    & 0     & 0 & 0 & 0 & 0 & 0    &      \\
129--256  & 500 & 0    & 0.04  & 0 & 0 & 0 & 0 & 1    & 22.36 \\
257--512  & 500 & 0    & 0     & 0 & 0 & 0 & 0 & 0    &      \\
513--1024 & 500 & 0.01 & 0.13  & 0 & 0 & 0 & 0 & 2    & 15.80 \\
1025--2048 & 500 & 0.12 & 1.84 & 0 & 0 & 0 & 0 & 38   & 15.06 \\
2049--4096 & 415 & 0.17 & 2.41 & 0 & 0 & 0 & 0 & 48   & 14.30 \\
4097+      & 134 & 1.88 & 19.48 & 0 & 0 & 0 & 0 & 225  & 10.36 \\
\bottomrule
\end{tabular}
}%
\caption{Infini-gram $n$-gram count statistics across citation buckets using RedPajama corpus (camelCase title + first author).}
\label{tab:infini-gram-redpajama-camel}
\end{table}

\section{More Details on Methodology}
\subsection{Data Sampling Strategy\label{sec:appendix_data_sampling}}

From the 1,921,209 papers in our curated dataset (Section~\ref{sec:dataset}), we drew a stratified sample of 9,108 papers to systematically test the relationship between factual prevalence (proxied by citation count) and hallucination rates. This section describes our two-dimensional stratification approach and the rationale for our sampling decisions.

\subsubsection{Stratification Dimensions}

\textbf{Field Selection:} We selected the 8 most prevalent disciplines from our initial sample: Medicine, Biology, Chemistry, Computer Science, Psychology, Physics, Materials Science, and Mathematics. These fields account for 77\% of papers in the full dataset and span diverse scientific domains with varying authorship conventions and publication patterns. This selection ensures sufficient sample sizes across citation bins while maintaining representativeness of the broader academic landscape.

\noindent \textbf{Citation Binning:} We defined 13 exponentially-spaced citation bins to capture the heavily right-skewed distribution of academic citations (Table~\ref{tab:citations}): [0--2], [3--4], [5--8], [9--16], [17--32], [33--64], [65--128], [129--256], [257--512], [513--1024], [1025--2048], [2049--4096], and [4097+]. This binning strategy serves three purposes:

\begin{itemize}[
   leftmargin=1em,
    itemsep=0.2em,
    topsep=0.2em,
    partopsep=0em,
    parsep=0em
]
\item \textbf{Fine granularity in low-citation ranges:} Since 36.9\% of papers have 0--2 citations, and 78.6\% have fewer than 33 citations, narrow bins in this range capture nuanced patterns where most papers reside.

\item \textbf{Sufficient representation in high-citation ranges:} Exponential spacing ensures adequate samples in the tail of the distribution (513+ citations), where highly-cited papers are naturally rare but hallucination patterns differ substantially.

\item \textbf{Computational efficiency:} Linear binning would require thousands of papers to adequately sample the tail, while exponential spacing achieves a more balanced representation with manageable sample sizes.
\end{itemize}

\subsubsection{Sampling Procedure and Coverage}

Our target was 100 papers per field per citation bin, yielding a theoretical maximum of $100 \times 8 \times 13 = 10,400$ papers. We employed stratified random sampling within each (field, bin) combination, selecting papers uniformly at random from the available pool.

Due to insufficient papers in some high-citation bins for certain fields, particularly Mathematics in bins above 2,049 citations, our final dataset comprises \textbf{9,108 papers}.

\subsection{LLM Prompt Design for Authorship Attribution}
\label{sec:appendix_prompt_design}

Figure~\ref{fig:prompt_verbose} is the complete verbatim prompt template for LLMs to generate author names, including six few-shot examples.

\begin{figure*}[h]
\centering
\begin{tcolorbox}[colback=gray!5, colframe=gray!80, boxrule=0.5pt, arc=2mm, title=\textbf{LLM Prompt for Generating Author Names}]
\small
\begin{verbatim}
You are an academic assistant. Your task is to answer questions about
the authors of academic papers. ONLY return a comma-separated list of
author names in the following format:
- Always put the last name in the end of each author name.
- Only return the author names.
- Do not add any explanation, notes, or extra content. Only generate
  the answer to the question.

Here are some examples:

Question: Who are the authors of the paper titled 'The relationship
between liver-kidney impairment and viral load after nephropathogenic
infectious bronchitis virus infection in embryonic chickens' which was
published in 2017?
Answer: Qiqi Huang, Xiaona Gao, Ping Liu, Huayuan Lin, Weilian Liu,
Guohui Liu, Jin Zhang, Guangfu Deng, Caiying Zhang, H. Cao, Xiaoquan
Guo, G. Hu

Question: Who are the authors of the paper titled 'Why men are at
higher risk for hepatocellular carcinoma?' which was published in 2012?
Answer: V. Keng, D. Largaespada, A. Villanueva

Question: Who are the authors of the paper titled 'Deciding laparoscopic
approaches for wedge resection in gastric submucosal tumors: a suggestive
flow chart using three major determinants' which was published in 2012?
Answer: Chung-Ho Lee, M. Hyun, Ye-Ji Kwon, S. Cho, Sung-Soo Park

Question: Who are the authors of the paper titled 'Use of internal and
external sources of knowledge and innovation in the Canadian wine
industry' which was published in 2015?
Answer: David Doloreux

Question: Who are the authors of the paper titled 'Dual phase helical CT
versus portal venous phase CT for the detection of colorectal liver
metastases: correlation with intra-operative sonography, surgical and
pathological findings' which was published in 2001?
Answer: D.John Scott, J. Guthrie, Paul Arnold, J. Ward, Julian Atchley,
Daniel J. Wilson, Philip J. Robinson

Question: Who are the authors of the paper titled 'Research as an
influence on teaching' which was published in 1984?
Answer: K. R. Jolls

Now, answer the following question:
Question: Who are the authors of the paper titled '{title}' which was published in '{year}'?
Answer:
\end{verbatim}
\end{tcolorbox}
\caption{LLM prompt for generating author names.}
\label{fig:prompt_verbose}
\end{figure*}

\subsubsection{Prompt Validation}

We validated the prompt's effectiveness through pilot testing on 100 randomly selected papers, examining:
\begin{itemize}[
   leftmargin=1em,
    itemsep=0.2em,
    topsep=0.2em,
    partopsep=0em,
    parsep=0em
]
\item \textbf{Format compliance:} 99\% of responses adhered to output format without explanatory text
\item \textbf{Name format accuracy:} 97\% correctly abbreviated first/middle names and preserved full last names
\end{itemize}

The few-shot examples proved essential: removing them resulted in a 35\% increase in format violations, confirming that in-context learning is necessary for consistent structured output.

\subsection{Hallucination Rate Definitions}
\label{sec:appendix_HR_definitions}

\subsubsection{Hallucination Metric Behavior Examples}

Table~\ref{tab:hr_example} illustrates how the three metrics behave across different error scenarios. Several patterns emerge:

\begin{itemize}[
   leftmargin=1em,
    itemsep=0.2em,
    topsep=0.2em,
    partopsep=0em,
    parsep=0em
]
\item \textbf{Perfect match:} All metrics return 0 when $M > 0$ and $H = U = 0$.
\item \textbf{One-sided errors:} HR$_3$ returns 0 when errors occur in only one direction (only hallucinations or only omissions), while HR$_1$ and HR$_2$ capture the magnitude of error.
\item \textbf{Severe confusion:} When both $H$ and $U$ are large, HR$_1$ and HR$_3$ approach HR$_2$, indicating the model is fundamentally confused about the paper's authorship.
\item \textbf{No overlap:} All metrics return 1.0 when $M = 0$ (complete failure).
\end{itemize}

\begin{table*}[t]
\centering

\begin{tabular}{@{}lcccccc@{}}
\toprule
 & \textbf{Matches} & \textbf{Hallucinated} & \textbf{Unretrieved} & & & \\
Scenario & $M$ & $H$ & $U$ & HR$_1$ & HR$_2$ & HR$_3$ \\
\midrule
Perfect match & 5 & 0 & 0 & 0.000 & 0.000 & 0.000 \\
Only hallucinations & 3 & 5 & 0 & 0.625 & 0.625 & 0.000 \\
Only omissions & 3 & 0 & 2 & 0.400 & 0.400 & 0.000 \\
Balanced errors & 3 & 2 & 2 & 0.400 & 0.571 & 0.400 \\
Severe confusion & 2 & 5 & 5 & 0.714 & 0.833 & 0.714 \\
Complete failure & 0 & 5 & 5 & 1.000 & 1.000 & 1.000 \\
\bottomrule
\end{tabular}
\caption{Example hallucination rate calculations across different error scenarios.}
\label{tab:hr_example}
\end{table*}

\subsubsection{Robustness Across Metrics}
\label{sec:robustness}

To verify that the findings of Section~\ref{sec:results} are not artifacts of metric choice, we examined consistency across HR$_1$, HR$_2$, and HR$_3$. Table~\ref{tab:metric_correlations} shows all three metrics are extremely highly correlated ($r > 0.98$, all $p < 10^{-10}$).

\begin{table}[htbp]
  \centering

  \begin{tabular}{@{}lrrr@{}}
    \toprule
    & \textbf{HR}$_1$ & \textbf{HR}$_2$ & \textbf{HR}$_3$ \\
    \midrule
    HR$_1$ & 1.000 & — & — \\
    HR$_2$ & 0.997 & 1.000 & — \\
    HR$_3$ & 0.982 & 0.984 & 1.000 \\
    \bottomrule
  \end{tabular}
    \caption{Pearson correlations between hallucination metrics (GPT-4o, $n = 9{,}108$ papers). All $p < 10^{-10}$.}
  \label{tab:metric_correlations}
\end{table}

Rank orderings across the 9,108 papers are nearly identical regardless of metric (Kendall's $\tau > 0.95$ for all pairwise comparisons). The citation-hallucination correlation holds with consistent magnitude across metrics: for Computer Science, $\rho_{\text{HR}_1} = -0.478$, $\rho_{\text{HR}_2} = -0.476$, $\rho_{\text{HR}_3} = -0.471$ (all $p < 0.001$). Field rankings remain identical across all three metrics. These results confirm that the central finding, the inverse relationship between citation prevalence and hallucination rate, is robust to metric specification.

\subsubsection{Metric Relationships and Selection}

As demonstrated in Table~\ref{tab:metric_correlations}, all three metrics are extremely highly correlated ($r > 0.98$) across our 9,108-paper dataset, confirming they capture the same underlying phenomenon. However, we report HR$_2$ as our primary metric for three reasons:

\begin{itemize}[
   leftmargin=1em,
    itemsep=0.2em,
    topsep=0.2em,
    partopsep=0em,
    parsep=0em
]
\item \textbf{Symmetric treatment:} HR$_2$ weights hallucinations and omissions equally, avoiding bias toward one error type.
\item \textbf{Intuitive interpretation:} HR$_2 = 0.80$ directly means ``80\% of unique authors were errors.''
\item \textbf{Mathematical elegance:} HR$_2$ is the Jaccard distance, a well-established similarity measure with known properties.
\end{itemize}

We report HR$_1$ and HR$_3$ results in Appendix~\ref{sec:details_results} to demonstrate robustness, while all main figures and correlation analyses are based on HR$_2$.

\subsubsection{Decomposing HR\texorpdfstring{$_2$}{2} into Hallucination and Omission Components}
\label{sec:appendix_hr2_decomposition}

We defined HR$_2$ (Equation~\ref{eqn:hr2}), as $(|H|+|U|) / (|M|+|H|+|U|)$. While this metric provides a comprehensive measure of overall prediction error, it combines two distinct failure modes: the fabrication of incorrect authors ($H$) and the omission of ground-truth authors ($U$). To ensure that the observed inverse relationship between citation count and HR$_2$ is not masking divergent trends between these modes, we decompose the error into a \textbf{Hallucination-Only Rate (HR$_4$)} and an \textbf{Omission-Only Rate (HR$_5$)}.

\textbf{Hallucination-Only Rate (HR$_4$):} The fraction of the LLM's output that consists of fabricated names.
\begin{equation}
    \text{HR}_4 = \frac{|H|}{|M| + |H|}\label{eqn:hr4}
\end{equation}

\textbf{Omission-Only Rate (HR$_5$):} The fraction of the ground-truth author set that the model failed to retrieve.
\begin{equation}
    \text{HR}_5 = \frac{|U|}{|M| + |U|}\label{eqn:hr5}
\end{equation}

As shown in Table~\ref{tab:spearman_hr}, both metrics exhibit nearly identical inverse correlations with citation count across all three models (all $p < 0.001$). The Spearman $\rho$ differences between HR$_4$ and HR$_5$ are below 0.01 in every case, suggesting that factual prevalence affects fabrication and recall in equal measure.

\begin{table}[tb!]
  \centering
  \resizebox{\columnwidth}{!}{%
  \begin{tabular}{@{}lccc@{}}
    \toprule
    \textbf{LLM} & \textbf{HR$_2$ (Total)} & \textbf{HR$_4$ (Halluc.)} & \textbf{HR$_5$ (Omiss.)} \\
    \midrule
    GPT-4o              & -0.385 & -0.385 & -0.383 \\
    DeepSeek-R1         & -0.472 & -0.475 & -0.469 \\
    Claude-Sonnet-4.5   & -0.462 & -0.461 & -0.458 \\
    \bottomrule
  \end{tabular}%
  }
  \caption{Spearman $\rho$ between citation count and each hallucination metric across all three models. All $p < 0.001$.}
  \label{tab:spearman_hr}
\end{table}

Furthermore, both components correlate extremely strongly with the composite HR$_2$ metric (Pearson $r > 0.94$), confirming that fabrication and omission co-occur proportionally as prevalence decreases rather than diverging (Table~\ref{tab:pearson_hr}).

\begin{table}[tb!]
  \centering
  \small

  \begin{tabular}{@{}lcc@{}}
    \toprule
    \textbf{LLM} & \textbf{HR$_4$ vs.\ HR$_2$} & \textbf{HR$_5$ vs.\ HR$_2$} \\
    \midrule
    GPT-4o              & 0.966 & 0.947 \\
    DeepSeek-R1         & 0.971 & 0.940 \\
    Claude-Sonnet-4.5   & 0.974 & 0.962 \\
    \bottomrule
  \end{tabular}
  \caption{Pearson $r$ between each component metric and the composite HR$_2$ metric.}
  \label{tab:pearson_hr}
\end{table}

To directly verify that these failure modes track together throughout the citation spectrum, we computed the absolute difference between HR$_4$ and HR$_5$ at each citation bin (averaged across all three models). The mean difference is 0.007, and the maximum is 0.022, confirming that hallucination and omission rates remain nearly identical across the full citation spectrum (Table~\ref{tab:hr4_hr5_binwise}).

\begin{table}[tb!]
  \centering
  \setlength{\tabcolsep}{4pt}
  \begin{tabular}{@{}lccc@{}}
    \toprule
    \textbf{Citation Bin} & \textbf{HR$_4$} & \textbf{HR$_5$} & $|\Delta|$ \\
    \midrule
0--2        & 0.981 & 0.978 & 0.003 \\
3--4        & 0.970 & 0.968 & 0.002 \\
5--8        & 0.975 & 0.972 & 0.004 \\
9--16       & 0.954 & 0.951 & 0.003 \\
17--32      & 0.942 & 0.937 & 0.005 \\
33--64      & 0.932 & 0.923 & 0.009 \\
65--128     & 0.897 & 0.896 & 0.001 \\
129--256    & 0.855 & 0.854 & 0.001 \\
257--512    & 0.795 & 0.784 & 0.011 \\
513--1024   & 0.702 & 0.680 & 0.022 \\
1025--2048  & 0.580 & 0.566 & 0.014 \\
2049--4096  & 0.482 & 0.471 & 0.012 \\
4097+       & 0.239 & 0.243 & 0.004 \\
    \bottomrule
  \end{tabular}
  \caption{HR$_4$ vs.\ HR$_5$ per citation bin, averaged across all three models. $|\Delta|$ denotes the absolute difference between the two rates.}
  \label{tab:hr4_hr5_binwise}
\end{table}

These results demonstrate that low-prevalence papers trigger a general state of uncertainty affecting all aspects of authorship recall: models simultaneously invent incorrect authors and forget real ones at nearly equal rates. HR$_2$ faithfully captures both failure modes without masking divergent trends.

\subsection{Evaluation Method Comparison}
\label{sec:appendix_eval_comparison}

To validate our evaluation approach, we compared two methods for assessing LLM-generated author attributions: \textbf{LLM self-evaluation} and \textbf{hard-coded evaluation}. This comparison serves both to validate our chosen evaluation strategy and to demonstrate the reliability of hard-coded evaluation for this task.

\subsubsection{Evaluation Methods}

\textbf{LLM self-evaluation} employs a structured prompt (Figure~\ref{fig:prompt_evaluation}) that asks the same LLM to evaluate its own output by comparing generated author names against the ground truth reference list. The evaluation prompt instructs the LLM to categorize each response and output three integer values: (X) the number of correctly matched authors, (Y) the number of hallucinated authors not present in the ground truth, and (Z) the number of ground truth authors missed by the LLM. These quantities directly correspond to $M$ (Matches), $H$ (Hallucinations), and $U$ (Unretrieved), as defined in Section~\ref{sec:hr_definitions}. The LLM performs fuzzy matching internally, accounting for variations in name formatting, ordering, and representation.

\begin{figure*}[htbp]
\centering
\begin{tcolorbox}[colback=gray!5, colframe=gray!80, boxrule=0.5pt, arc=2mm, title=\textbf{LLM Prompt for Author Matching Evaluation}]
\small
\begin{verbatim}
You are a research assistant evaluating an answer to a question about
authors of an academic paper.

Question: {question}
Reference answer (ground truth): {reference_answer}
Answer (to be evaluated): {generated_answer}

Both the answer and the reference answer listed authors separated by
commas.

Compare the answer with the reference answer, and give the following
three numbers in the format X, Y, Z where:
X is the number of authors in the answer that are also in the reference
  answer
Y is the number of authors in the answer but are not in the reference
  answer
Z is the number of authors in the reference answer but are not in the
  answer

Additional note:
- Cases like 'I. V. Serban' vs 'IV Serban' should be considered matching.
- Cases like 'I Serban' vs 'IV Serban' should be considered matching.
- Cases like 'IA Serban' vs 'IV Serban' should not be considered matching.
- Cases like 'AV Serban' vs 'IV Serban' should not be considered matching.
- Ignore the order of names.

Respond ONLY with three numbers.
For example, if an answer lists five authors, two appear in the references
while three do not, and there are four additional authors in the references
not listed in the answer, then your response should be: 2, 3, 4
\end{verbatim}
\end{tcolorbox}
\caption{LLM prompt for author matching evaluation.}
\label{fig:prompt_evaluation}
\end{figure*}

\textbf{Hard-coded evaluation} employs deterministic algorithmic fuzzy string matching to compare LLM-generated author names against ground truth, as mentioned in Section~\ref{sec:eval_comparison}.
The algorithm produces the same three metrics (X, Y, Z) as LLM self-evaluation, enabling direct comparison.

For the agreement analysis, as well as the variance and stability analyses presented in later sections, we use the responses generated by GPT-4o to compare the two evaluation methods.

\subsubsection{Agreement Analysis}

\begin{table}[htbp]
\setlength{\tabcolsep}{2.5pt}
  \centering
\resizebox{\columnwidth}{!}{%

  \begin{tabular}{@{}lcccc@{}}
    \toprule
    \textbf{Metric} & \textbf{Pearson} $r$ & \textbf{RMSE} & \textbf{MAE} & \textbf{Agreement $<$ 0.1} \\
    \midrule
    HR$_1$ & 0.970 & 0.065 & 0.009 & 97.1\% \\
    HR$_2$ & 0.969 & 0.067 & 0.010 & 96.9\% \\
    HR$_3$ & 0.963 & 0.088 & 0.014 & 95.8\% \\
    \bottomrule
  \end{tabular}%
}
    \caption{Agreement Between LLM Self-Evaluation and Hard-coded Evaluation. Agreement: \% of papers with absolute HR difference $<$ 0.1 between eval methods.}
  \label{tab:eval_agreement}

\end{table}

We applied both evaluation methods to all responses generated by GPT-4o, computing hallucination rates using both approaches. Table~\ref{tab:eval_agreement} presents the overall agreement metrics across all three hallucination rate definitions.

The two methods demonstrate strong agreement across all metrics. For our primary metric (HR$_2$, Jaccard Hallucination Rate), we observe a Pearson correlation of $r = 0.969$ ($p < 0.001$), with a root mean squared error of 0.067 and a mean absolute error of 0.010. Notably, 96.9\% of papers exhibit an absolute HR difference below 0.1, indicating that the two methods produce practically equivalent results for the vast majority of cases. The slightly lower correlation for HR$_3$ ($r = 0.963$) reflects the increased sensitivity of this metric to edge cases where hallucinations dominate the response.

\subsubsection{Variance and Stability of Hallucination Metrics\label{sec:appendix_HRs}}

Table~\ref{tab:variance_comparison} compares the variance of hallucination rates produced by each method. LLM self-evaluation exhibits 1.04-1.05$\times$ higher variance across all three metrics, indicating slightly less consistency in repeated evaluations. This difference, while small, reflects the stochastic nature of LLM inference: identical inputs can produce subtly different evaluation outputs due to temperature sampling and other generation parameters. In contrast, hard-coded evaluation produces identical results for identical inputs, ensuring perfect reproducibility.

\begin{table}[htbp]
  \centering
  \normalsize

  \begin{tabular}{@{}lrrr@{}}
    \toprule
    \textbf{Metric} & \textbf{Var(LLM)} & \textbf{Var(Code)} & \textbf{Ratio} \\
    \midrule
    HR$_1$   & 0.0649 & 0.0623 & 1.04 \\
    HR$_2$     & 0.0600 & 0.0570 & 1.05 \\
    HR$_3$     & 0.0867 & 0.0834 & 1.04 \\
    \bottomrule
  \end{tabular}
    \caption{Variance comparison between evaluation methods.}
  \label{tab:variance_comparison}
\end{table}

\subsubsection{Method Selection Rationale}

Despite the strong agreement between the two evaluation methods, we ultimately chose \textbf{hard-coded evaluation} as our primary method for the following reasons:

\begin{itemize}[
   leftmargin=1em,
    itemsep=0.2em,
    topsep=0.2em,
    partopsep=0em,
    parsep=0em
]
\item \textbf{Perfect Reproducibility:} Hard-coded evaluation is deterministic: identical inputs always produce identical outputs with zero variance. This property is critical for scientific reproducibility and enables other researchers to verify our results exactly.

\item \textbf{Transparency and Auditability:} The matching algorithm is fully transparent and can be inspected, debugged, and validated line-by-line. When disagreements occur, we can trace the exact cause. LLM self-evaluation operates as a black box, making it difficult to understand or debug edge cases.

\item \textbf{Zero Computational Cost:} Hard-coded evaluation requires no API calls, eliminating both monetary costs and dependencies on external service availability. For a dataset of 9,108 papers, this difference is substantial.

\item \textbf{Model Agnosticism:} Hard-coded evaluation works identically regardless of which LLM generated the attribution, enabling fair cross-model comparisons without the confound of having different models evaluate their own outputs differently.

\item \textbf{No Meta-Evaluation Concerns:} LLM self-evaluation introduces a meta-evaluation problem. We must trust the LLM to accurately assess its own performance, which may be subject to model-specific biases or inconsistencies in the interpretation of evaluation criteria.
\end{itemize}

Taken together, hard-coded evaluation provides superior reproducibility, transparency, and practicality for large-scale analysis.

\section{Additional Analysis and Results}
\label{sec:details_results}

\subsection{Hallucination Rates Across Models and Fields}
\label{sec:appendix_results_hr_extra}

\begin{figure}[t]
  \centering
  \includegraphics[width=\columnwidth]{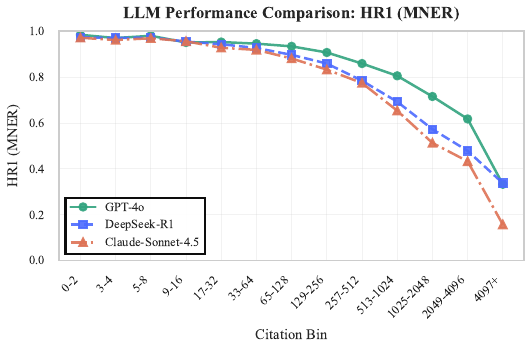}
  \caption{Hallucination rate (HR$_1$) across models as a function of citation count.}
  \label{fig:llm_comparison_hr1}
\end{figure}

\begin{figure}[t]
  \centering
  \includegraphics[width=\columnwidth]{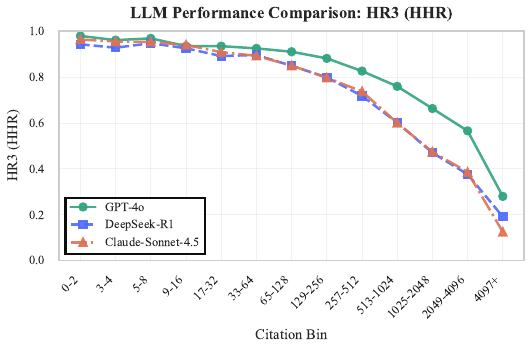}
  \caption{Hallucination rate (HR$_3$) across models as a function of citation count.}
  \label{fig:llm_comparison_hr3}
\end{figure}

Figure~\ref{fig:llm_comparison_hr1} and Figure~\ref{fig:llm_comparison_hr3} plot hallucination rates under HR$_1$ and HR$_3$ as functions of citation count. Consistent patterns are observed across all three LLMs: hallucination rates decline monotonically as citation count increases, closely mirroring the trend reported for HR$_2$ in Figure~\ref{fig:hr_by_citations}. For low-citation papers, both metrics yield near-unity hallucination rates, indicating substantial difficulty in author attribution when factual prevalence is minimal. As citation count increases, performance improves steadily across models, reinforcing the existence of a systematic relationship between citation prevalence and author recall.

Compared with HR$_2$, HR$_1$ exhibits trends highly similar to those of HR$_2$, whereas HR$_3$ yields systematically lower values because it penalizes only cases where hallucinations and omissions co-occur, while cases involving only one type of error are barely penalized. Despite these scale differences, the relative ordering of models and the citation-dependent trends remain stable. These results further validate that the main findings based on HR$_2$ are robust to alternative hallucination rate definitions.

Table~\ref{tab:multi_llm_summary} reports the overall hallucination rates (mean $\pm$ standard deviation) for three LLMs under the three hallucination rate definitions (HR$_1$, HR$_2$, and HR$_3$). Across all models, the three metrics exhibit consistent relative ordering, with HR$_1$ yielding the highest average hallucination rates and HR$_3$ the lowest. This pattern reflects the different treatment of partial matches across the definitions. While absolute values differ, the results show that the choice of hallucination rate definition primarily affects scale rather than the relative comparison between models, with GPT-4o consistently exhibiting higher hallucination rates than DeepSeek-R1 and Claude Sonnet 4.5 under all three definitions. The relatively large standard deviations indicate substantial variability at the paper level, consistent with the heterogeneous difficulty of author attribution across citation levels.

\begin{table}[htbp]
  \centering
  \normalsize

  \resizebox{\columnwidth}{!}{%
  \begin{tabular}{@{}lccc@{}}
    \toprule
    \textbf{Model} & \textbf{HR$_1$} & \textbf{HR$_2$} & \textbf{HR$_3$} \\
    \midrule
    GPT-4o & 0.90{\small$\pm$0.25} & 0.91{\small$\pm$0.24} & 0.87{\small$\pm$0.29} \\
    DeepSeek-R1 & 0.85{\small$\pm$0.29} & 0.87{\small$\pm$0.28} & 0.80{\small$\pm$0.36} \\
    Claude-Sonnet-4.5 & 0.83{\small$\pm$0.32} & 0.85{\small$\pm$0.32} & 0.81{\small$\pm$0.36} \\
    \bottomrule
  \end{tabular}
}%
  \caption{Overall Hallucination Rate (Mean $\pm$ SD) by LLM.}
  \label{tab:multi_llm_summary}
\end{table}

Table~\ref{tab:detailed_gpt_4o} presents a field-level breakdown of GPT-4o performance using HR$_2$. Fields are sorted by mean HR$_2$, revealing pronounced domain differences. Materials Science and Chemistry exhibit the highest hallucination rates, whereas Computer Science and Physics show substantially lower values. The standard deviation $\sigma$ indicates increasing performance variability in lower-HR fields, suggesting a wider spread of attribution difficulty within those domains. The Spearman correlation coefficients $\rho$ are negative across all fields and statistically significant, confirming a consistent inverse relationship between citation count and hallucination rate at the domain level.

\begin{table}[htbp]
\setlength{\tabcolsep}{3pt}
  \centering

  \begin{tabular}{@{}lrrrrr@{}}
    \toprule
    \textbf{Field} & \textbf{Pap.} & \textbf{Aut.} & \textbf{HR$_2$} & $\sigma$ & $\rho$ \\
    \midrule
    Materials Sci. & 1,067 & 4.2 & 0.971 & 0.132 & -0.228 \\
    Chemistry & 1,134 & 3.8 & 0.955 & 0.159 & -0.323 \\
    Biology & 1,179 & 4.6 & 0.927 & 0.194 & -0.400 \\
    Psychology & 1,154 & 2.9 & 0.907 & 0.243 & -0.409 \\
    Medicine & 1,230 & 5.0 & 0.906 & 0.231 & -0.453 \\
    Mathematics & 1,088 & 2.2 & 0.901 & 0.258 & -0.364 \\
    Computer Sci. & 1,174 & 3.2 & 0.874 & 0.298 & -0.476 \\
    Physics & 1,082 & 3.7 & 0.850 & 0.304 & -0.383 \\
    \midrule
    \textbf{Overall} & \textbf{9,108} & \textbf{3.7} & \textbf{0.911} & \textbf{0.237} & \textbf{-0.385} \\
    \bottomrule
  \end{tabular}
    \caption{GPT-4o performance by field (sorted by HR$_2$). Pap.: number of papers; Aut.: mean number of authors per paper; $\rho$: Spearman correlation between citation count and HR$_2$; $\sigma$: standard deviation of HR$_2$ within each field. All $p < 0.001$.}
  \label{tab:detailed_gpt_4o}
\end{table}

\subsection{Author Position and Hallucination Rate}
\label{sec:author_position_hallucination}

In multi-author papers, author order reflects contribution, with first and last authors typically most prominent. We examine whether LLM recall accuracy varies by author position, restricting analysis to responses where at least one author is correctly identified to ensure minimal knowledge of the cited work.

\begin{figure}[htbp]
   \centering
   \hspace*{-0.14cm}
    \begin{minipage}{0.485\columnwidth}
        \centering
        \includegraphics[width=\linewidth]{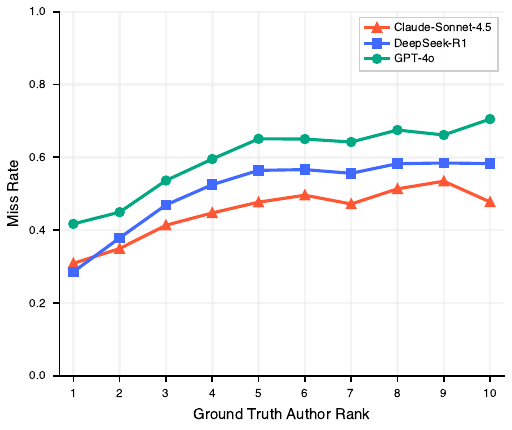}
 \vspace{-0.48cm}
        \caption*{\scriptsize (a) Miss Rate by Ground Truth \\ Position}
    \end{minipage}%
    \begin{minipage}{0.485\columnwidth}
        \centering
        \includegraphics[width=\linewidth]{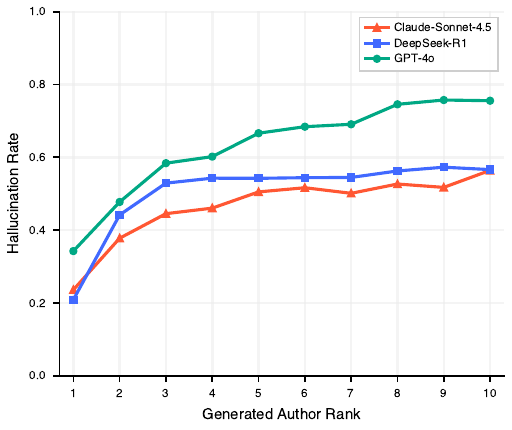}
        \captionsetup{justification=centering}
        \caption*{\hspace{0.1cm} \scriptsize  (b) Hallucination Rate by Generated \\ Position}
    \end{minipage}
 \vspace{-0.2cm}
 \captionsetup{font=small}
    \caption{Positional bias in authorship hallucination.
(a) Failure to retrieve the $n$-th author increases with rank.
(b) Hallucination likelihood for the $n$-th generated name rises as the list length increases.}

    \label{fig:position_bias}
\end{figure}

We found that performance is highest for early authors and declines rapidly across all models (Figure~\ref{fig:position_bias}). First authors are consistently recalled most accurately (e.g., GPT-4o miss rate 41.7\%), while error rates rise sharply by the 10th author (70.5\%), with no recall advantage for senior (last) authors. A similar trend holds for generation quality: hallucination rates are low for the first name (20.9\% DeepSeek-R1, 23.6\% Claude Sonnet, 34.2\% GPT-4o) but increase substantially by the tenth (56.6\%, 56.4\%, and 75.5\%), indicating that while models can often identify the primary contributor, they struggle to accurately recall the full research team.

\subsection{Author Count and Hallucination Rate\label{sec:author_count_hallucination}}
We examined whether the number of ground truth authors affects hallucination rates. Surprisingly, we found a small but significant \textit{negative} correlation: papers with more authors exhibited lower HR$_2$. To understand this counterintuitive result, we analyzed model output behavior.

Models tend to output multiple author names regardless of true author count. For single-author papers, models output 2.5 names on average, of which 2.3 are hallucinated, more than 2 fabricated names for every real author. For 5-author papers, models output 4.8 names with 4.1 hallucinated, a similar total but less than 1 fabricated name per real author. This explains the negative correlation: papers with fewer authors accumulate more hallucinations relative to their size, inflating their error rates. Table~\ref{tab:author_output} shows this pattern across all author counts.

\begin{table}[htbp]
\centering
\small

\begin{tabular}{lccc}
\toprule
\textbf{GT Authors} & \textbf{Output} & \textbf{Halluc.} & \textbf{Halluc/Author} \\
\midrule
1      & 2.5 & 2.3 & 2.3 \\
2--3   & 3.4 & 3.0 & 1.2 \\
4--5   & 4.5 & 3.9 & 0.9 \\
6--10  & 6.5 & 5.5 & 0.7 \\
11+    & 10.0 & 7.0 & 0.4 \\
\bottomrule
\end{tabular}
\caption{LLM output behavior by ground truth author count, averaged across models. Output: typical author count produced by the LLM. Halluc.: average hallucinated name count. Halluc/Author = hallucinated names per true author.}
\label{tab:author_output}
\end{table}

This effect varies by field (Table~\ref{tab:field_author}). Models appear to learn field-specific expectations about author counts. Even for single-author papers, models output 3.1 names in medicine (where multi-author papers are typical) versus only 2.1 in mathematics (where single-author papers are common). Since more output means more chances to match true authors, this field-aware behavior results in stronger negative correlations in high-author fields like medicine ($\rho = -0.16$) compared to mathematics ($\rho = -0.05$).

\begin{table}[htbp]
\centering
\small

\begin{tabular}{lcccc}
\toprule
 & \textbf{Typical} & \textbf{Output} & \textbf{Halluc.} & \\
\textbf{Field} & \textbf{Authors} & \textbf{(GT=1)} & \textbf{(GT=1)} & \textbf{$\rho$} \\
\midrule
Medicine      & 4.94 & 3.1 & 2.9 & -0.16 \\
Biology       & 4.49 & 2.8 & 2.6 & -0.11 \\
Physics       & 3.72 & 2.5 & 2.2 & -0.11 \\
Psychology    & 2.85 & 2.4 & 2.3 & -0.13 \\
Mathematics   & 2.20 & 2.1 & 1.9 & -0.05 \\
\bottomrule
\end{tabular}
\caption{Output behavior by field, averaged across models. Typical authors: mean author count in the field. Output and Halluc.: model behavior for single-author papers. $\rho$: correlation between author count and HR$_2$.}
\label{tab:field_author}
\end{table}

In summary, models infer author count from context: their output scales with the actual number of authors ($\rho \approx 0.55$) and adjusts by field. The negative correlation between author count and HR$_2$ arises because models over-produce names for papers with 1--2 authors (inflating their errors), while papers with more authors have more chances for correct matches (lowering their error rate). Thus, author count interacts with model output patterns to influence measured error rates.

\subsection{Ethnic Distribution of Hallucinated Names
\label{sec:ethnicity}}
We investigated whether LLMs exhibit systematic bias in the ethnic composition of hallucinated author names. Using \texttt{ethnicseer}, a name-based ethnicity classifier, we categorized all unique last names from our 1.9M academic paper dataset (715,837 names) and all hallucinated names from the three models into 12 ethnic categories: English/Western, Chinese, German, Japanese, Indian, Italian, French, Spanish, Russian, Middle Eastern, Korean, and Vietnamese.

We conducted two complementary analyses. The \textbf{population analysis} counts each hallucinated name by how many unique papers it appeared in, measuring the ethnic breakdown of all hallucination instances a user would encounter. The \textbf{diversity analysis} counts each unique hallucinated name once, regardless of frequency, measuring the variety of names the model can produce for each ethnic group. For both analyses, we calculated the ratio of each ethnicity's share among hallucinated names to its share in the dataset. A ratio of 1.0 indicates no bias, while values above or below indicate over- or under-representation.

To test for bias, we compared the ethnic distribution of hallucinated names against the dataset baseline using chi-square goodness-of-fit tests with Bonferroni correction ($\alpha' = 0.0167$). Under the null hypothesis, the ethnic proportions of hallucinated names should match those in the academic dataset.

\begin{figure}[htbp]
\centering
\includegraphics[width=\columnwidth]{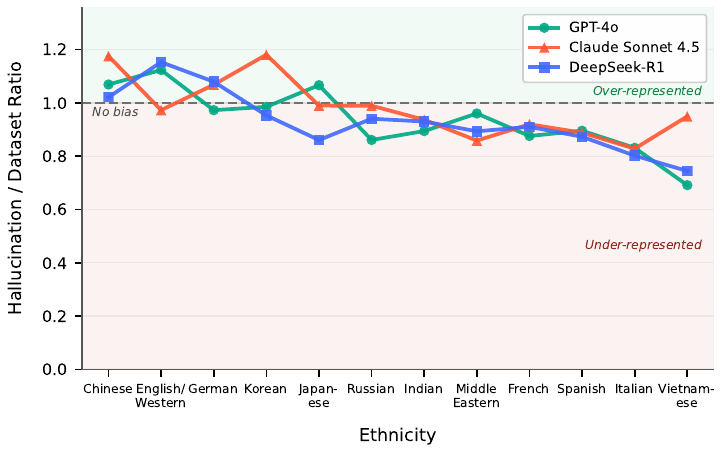}
\caption{Population analysis: ratio of each ethnicity's share among hallucinated name instances to its share in the dataset. Values near 1.0 indicate no bias. All models closely match the dataset baseline.}
\label{fig:ethnicity_population}
\end{figure}

When counting by paper frequency, all three models showed statistically significant but \textit{practically negligible} deviation from the baseline (Table~\ref{tab:ethnicity_bias}). Cramér's $V$ ranged from 0.031 to 0.035, indicating that the ethnic distribution of hallucinated name instances closely mirrors the dataset. For example, Chinese names constitute 16.2\% of name occurrences in the dataset and 17.3\% of hallucination instances, a difference of just 1.1 percentage points. In other words, if a user receives a hallucinated author name, it's likely ethnicity roughly matches what would be expected from the underlying academic population (Figure~\ref{fig:ethnicity_population}).

\begin{figure}[t]
\centering
\includegraphics[width=\columnwidth]{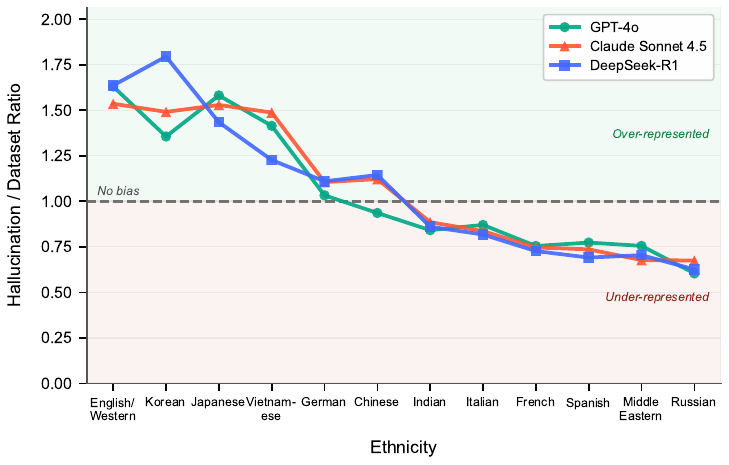}
\caption{Diversity analysis: ratio of each ethnicity's share among unique hallucinated names to its share in the dataset. English/Western and Japanese names are modestly over-represented, while Russian and French names are under-represented.}
\label{fig:ethnicity_diversity}
\end{figure}

When counting unique names, a slightly larger but still small effect emerged ($V = 0.09$--$0.10$). English/Western names were consistently over-represented across all models (ratio 1.5--1.6$\times$), as were Korean (1.4--1.8$\times$) and Japanese (1.4--1.6$\times$) names. DeepSeek showed notably higher over-representation of Korean names (1.79$\times$). Conversely, Russian (0.60--0.67$\times$) and French (0.73--0.75$\times$) names were consistently under-represented (Figure~\ref{fig:ethnicity_diversity}). For example, GPT-4o hallucinated 4.3\% of English/Western names in the dataset pool (4,762 of 110,279) compared to only 1.6\% of Russian names (1,028 of 64,251), a 2.7$\times$ difference in sampling rate.

\begin{table}[htbp]
\centering

\begin{tabular}{@{}lcc@{}}
\toprule
\textbf{Model} & \textbf{Population} & \textbf{Diversity} \\
\midrule
GPT-4o & 0.031 & 0.101 \\
Claude Sonnet 4.5 & 0.031 & 0.093 \\
DeepSeek-R1 & 0.035 & 0.103 \\
\bottomrule
\end{tabular}
\caption{Cramér's $V$ effect sizes for ethnic bias in hallucinated names. All $p < 0.001$.}
\label{tab:ethnicity_bias}
\end{table}

These findings indicate that ethnic bias in LLM hallucinations, while statistically detectable, is practically modest. Both the \textit{population} of hallucinated names and their \textit{diversity} across ethnic groups largely mirror the academic dataset baseline, with only small deviations. The consistency of these patterns across three architecturally distinct models suggests they arise from shared properties of web-scale training data rather than model-specific artifacts.

\subsection{Title Length and Hallucination Rate
\label{sec:title_length}}

We examined whether paper title length correlates with hallucination rates, controlling for citation count. Longer titles often reflect more specialized or complex topics, which models may find harder to recall accurately. We measured title length in characters and computed Spearman correlations with all three hallucination metrics (HR$_1$, HR$_2$, HR$_3$).

To isolate the independent effect of title length from citation count, we used two approaches. First, we computed correlations separately within each of 13 citation buckets, then combined them using Fisher's Z transformation. This ensures papers are only compared against others with similar citation counts. Second, we computed partial correlations, which statistically remove any variance that title length shares with citation count.

\begin{table}[htbp]
\centering
\small

\begin{tabular}{lcc}
\toprule
\textbf{Model} & \textbf{Overall} & \textbf{Citation-Adjusted} \\
\midrule
GPT-4o            & + 0.035 &+ 0.007  \\
Claude Sonnet 4.5 & + 0.048 & + 0.017 \\
DeepSeek-R1       & + 0.050 & + 0.019 \\
\bottomrule
\end{tabular}
\caption{Spearman correlation ($\rho$) between title length and HR$_2$. Overall: unadjusted. Citation-Adjusted: partial correlation controlling for citation count.}
\label{tab:title_length}
\end{table}

As shown in Table~\ref{tab:title_length}, the unadjusted correlations were small but statistically significant ($\rho = +0.035$ to $+0.050$, $p < 0.001$). After controlling for citation count, all three fall to near zero ($\rho = +0.007$ to $+0.019$) and none remains significant ($p = 0.076$ to $0.479$). Title length therefore has no reliable effect of its own once citations are accounted for.

\subsection{Multivariate Regression and Mixed-Effects Analysis}
\label{sec:regression}

While our field-stratified analysis (Section~\ref{sec:results}, Table~\ref{tab:spearman_all_models}) partially controls for field-specific effects, and our supplementary analyses examine author count (Appendix~\ref{sec:author_count_hallucination}) and title length (Appendix~\ref{sec:title_length}) individually, these analyses do not jointly control for all potential confounders. To strengthen the argument that citation count has an independent effect on hallucination rates, we conduct multivariate regression analyses across multiple model specifications.

We fit generalized linear models (GLMs) with a binomial family and logit link to account for the bounded $[0,1]$ nature of hallucination rates, with $\log_2(\text{citation count} + 1)$, author count, title length, publication year, and field (dummy-coded with Biology as baseline) as independent variables, evaluated on both HR$_2$ (Equation~\ref{eqn:hr2}) and HR$_4$ (Equation~\ref{eqn:hr4}). Under both metrics, the effect of log-citation is negative, large, and highly significant (Table~\ref{tab:regression-summary}), with each doubling of citations reducing the odds of hallucination by 34--36\%. Notably, title length is not significant under either metric ($p > 0.37$), and author count becomes non-significant under HR$_4$ ($p = 0.463$), confirming that the citation--hallucination relationship is not an artifact of any particular covariate or evaluation choice.

To further control for unobserved field-level confounders, including naming conventions, transliteration practices, and collaboration norms, we fit a linear mixed-effects model with a random intercept per field ($k = 8$, $N = 9{,}108$), allowing each field its own baseline hallucination rate without requiring explicit enumeration of field-level factors. The fixed effect of log-citation remained highly significant and stable (Table~\ref{tab:regression-summary}), and the estimated intraclass correlation coefficient (ICC) was approximately 2\%, indicating that unobserved field-level factors account for only a negligible share of residual variance after controlling for citation count and other covariates.

\begin{table}[htbp]
\centering
\footnotesize
\resizebox{\columnwidth}{!}{%
\begin{tabular}{llccc}
\toprule
\textbf{Model} & \textbf{Metric} & $\boldsymbol{\beta}$ & $\boldsymbol{z}$ & $\boldsymbol{p}$ \\
\midrule
GLM, binomial logit & HR$_2$ & $-0.440$ & $-26.6$ & $<0.001$ \\
GLM, binomial logit & HR$_4$ & $-0.414$ & $-28.1$ & $<0.001$ \\
Mixed-effects LMM & HR$_2$ & $-0.030^{\dagger}$ & $-40.4$ & $<0.001$ \\
\bottomrule
\end{tabular}%
}
\caption{Citation coefficient across model specifications. $^{\dagger}$Coefficient on probability scale; GLM coefficients are on log-odds scale.}
\label{tab:regression-summary}
\end{table}

While direct measures of title ambiguity and author name transliteration would further enrich the analysis, the convergent evidence across metrics, model families, and the negligible ICC of 2\% strongly suggest that these unobserved factors do not materially confound our conclusions.

\subsection{Prompt Sensitivity Analysis}
\label{sec:appendix_prompt_sensitivity}

Because LLM outputs can be sensitive to prompt construction~\citep{zhao2021calibrate, lu2022fantastically}, we tested whether the observed relationship between factual prevalence and hallucination rates is sensitive to prompt wording. We conducted a sensitivity analysis using DeepSeek-R1 with three distinct prompt variants on a proportional stratified subsample ($n = 1{,}004$). This subsample was drawn across all 13 citation bins and 8 academic disciplines, preserving coverage across the citation and field strata used in the main evaluation.

Each variant modified the prompt text while preserving the core task structure and few-shot examples:

\begin{itemize}[
    leftmargin=1em,
    itemsep=0.2em,
    topsep=0.2em,
    partopsep=0em,
    parsep=0em
]
    \item \textbf{Baseline:} The original task-oriented prompt used for primary evaluation.
    \item \textbf{Emotional:} A variant emphasizing care and accuracy, framing the task as supporting a researcher's learning through deep attention to detail.
    \item \textbf{Authoritative:} A variant adopting a seasoned expert persona with scholarly authority, delivered with decisive precision.
\end{itemize}

All three prompt variants were queried using the same API settings: temperature~$=0$ and a 500-token response budget. Thus, this analysis varies only the prompt text, not stochastic decoding behavior.

As shown in Table~\ref{tab:spearman_prompt}, all three variants produced similar negative Spearman correlations between citation count and hallucination rate. To test whether these differences were statistically distinguishable from baseline, we aligned the three prompt variants by paper and ran paired nonparametric bootstrap tests over the same 1,004 records. For HR$_2$, the emotional prompt changed the citation--HR Spearman correlation by $\Delta\rho=-0.0260$ relative to baseline (95\% CI $[-0.0681, 0.0174]$, $p=0.233$), while the authoritative prompt changed it by $\Delta\rho=-0.0270$ (95\% CI $[-0.0718, 0.0154]$, $p=0.221$; Table~\ref{tab:prompt_spearman_diffs}). Across all HR metrics, the paired confidence intervals for Spearman differences included zero, indicating no statistically reliable change from baseline. Thus, the inverse citation--hallucination relationship remains stable across these prompt wordings.

\setlength{\tabcolsep}{0.5pt}
\begin{table}[tb!]
  \centering
  \captionsetup{font=small}
  \footnotesize
  \begin{tabular}{@{}lccccc@{}}
    \toprule
    \textbf{Prompt Variant} & \textbf{HR$_1$} & \textbf{HR$_2$} & \textbf{HR$_3$} & \textbf{HR$_4$} & \textbf{HR$_5$} \\
    \midrule
    Baseline      & $-0.447$ & $-0.450$ & $-0.446$ & $-0.453$ & $-0.450$ \\
    Emotional     & $-0.472$ & $-0.474$ & $-0.464$ & $-0.473$ & $-0.472$ \\
    Authoritative & $-0.473$ & $-0.474$ & $-0.470$ & $-0.478$ & $-0.470$ \\
    \midrule
    \textbf{Max Diff} & $\mathbf{0.026}$ & $\mathbf{0.025}$ & $\mathbf{0.024}$ & $\mathbf{0.025}$ & $\mathbf{0.022}$ \\
    \bottomrule
  \end{tabular}
  \caption{Spearman $\rho$ between citation count and each hallucination metric by prompt variant. All listed correlations have $p < 0.001$.}
  \label{tab:spearman_prompt}
\end{table}

\setlength{\tabcolsep}{3pt}
\begin{table}[tb!]
  \centering
  \captionsetup{font=small}
  \footnotesize
  \resizebox{\columnwidth}{!}{%
  \begin{tabular}{@{}lrrr@{}}
    \toprule
    \textbf{Comparison} & \textbf{$\Delta\rho$ for HR$_2$} & \textbf{95\% bootstrap CI} & \textbf{$p$} \\
    \midrule
    Emotional - Baseline & -0.0260 & $[-0.0681, 0.0174]$ & 0.233 \\
    Authoritative - Baseline & -0.0270 & $[-0.0718, 0.0154]$ & 0.221 \\
    \bottomrule
  \end{tabular}%
  }
  \caption{Paired bootstrap tests for prompt-variant Spearman differences. Each bootstrap sample resamples the same 1,004 papers across prompt variants, preserving the paired design. The $\Delta\rho$ values are estimated under this paired design and need not equal the difference of the marginal correlations reported in Table~\ref{tab:spearman_prompt}.}
  \label{tab:prompt_spearman_diffs}
\end{table}

Furthermore, the mean metric values across the entire subsample showed negligible shifts. The maximum absolute shift from the baseline across any metric was only 0.0139 (Table~\ref{tab:mean_prompt}). These results show that mean hallucination rates changed little across prompt variants.

\setlength{\tabcolsep}{0.5pt}
\begin{table}[tb!]
  \centering
  \captionsetup{font=small}
  \footnotesize

  \resizebox{\columnwidth}{!}{%
  \begin{tabular}{@{}lccccc@{}}
    \toprule
    \textbf{Prompt Variant} & \textbf{HR$_1$} & \textbf{HR$_2$} & \textbf{HR$_3$} & \textbf{HR$_4$} & \textbf{HR$_5$} \\
    \midrule
    \multicolumn{6}{@{}l}{\textit{Mean Values}} \\[2pt]
    Authoritative & $0.8565$ & $0.8678$ & $0.8144$ & $0.8451$ & $0.8240$ \\
    Baseline      & $0.8513$ & $0.8644$ & $0.8005$ & $0.8352$ & $0.8233$ \\
    Emotional     & $0.8500$ & $0.8619$ & $0.8022$ & $0.8358$ & $0.8156$ \\
    \midrule
    \multicolumn{6}{@{}l}{\textit{Absolute Difference from Baseline}} \\[2pt]
    Authoritative & $+0.0052$ & $+0.0034$ & $+0.0139$ & $+0.0099$ & $+0.0007$ \\
    Emotional     & $-0.0013$ & $-0.0025$ & $+0.0017$ & $+0.0006$ & $-0.0077$ \\
    \bottomrule
  \end{tabular}%
  }
  \caption{Mean metric values and absolute differences from baseline across different prompt variants.}
  \label{tab:mean_prompt}
\end{table}

Together, the similar Spearman correlations, nonsignificant paired bootstrap differences, and small mean metric shifts show that the long-tail hallucination pattern is robust to prompt variation.

\section{More Details on Multiple-choice Recognition Analysis}
\label{sec:appendix_mc_recognition}

This section provides the full experimental setup and results for the multiple-choice (MC) recognition experiment summarized  in Section~\ref{sec:mc_recognition}. The experiment is designed to test whether recall failures for low-prevalence papers reflect genuine knowledge absence or a retrieval limitation.

\subsection{Experimental Setup}

\noindent\textbf{Model and Data.}
We use GPT-4o (temperature~$= 0$) on the same 9{,}108 papers described in Section~\ref{sec:methodology}.

\noindent\textbf{Distractor Pool Construction.}
For each academic field, we collect all unique author names that appear in the evaluation set for that field (exact string deduplication). For a paper with $k$ true authors, we sample $3k$ candidate names without replacement from the corresponding field pool, excluding any candidate that fuzzy-matches a true author. The $3k$ names are then divided into three distractor groups of $k$ names each. We verified that across all 9{,}108 items, no distractor group contains duplicate names and no distractor name overlaps with the true author list.

\noindent\textbf{Choices Construction.}
Each MC item presents four options (A--D): the correct complete author list and three distractor lists, each containing exactly $k$ names. The position of the correct option is assigned uniformly at random (resulting distribution: A~=~24.3\%, B~=~25.4\%, C~=~24.4\%, D~=~26.0\%, all within $\pm 2$\,pp of uniform, indicating no systematic positional bias).

\noindent\textbf{Prompt Template Design.}
We adapt the few-shot authorship attribution prompt from Section~\ref{sec:prompt_design} to MC format. The model receives the paper title and publication year, followed by four labeled options (A--D), and is instructed to return a single letter. Two demonstration examples are included in context.

\begin{figure}[htbp]
  \centering
  \includegraphics[width=\columnwidth]{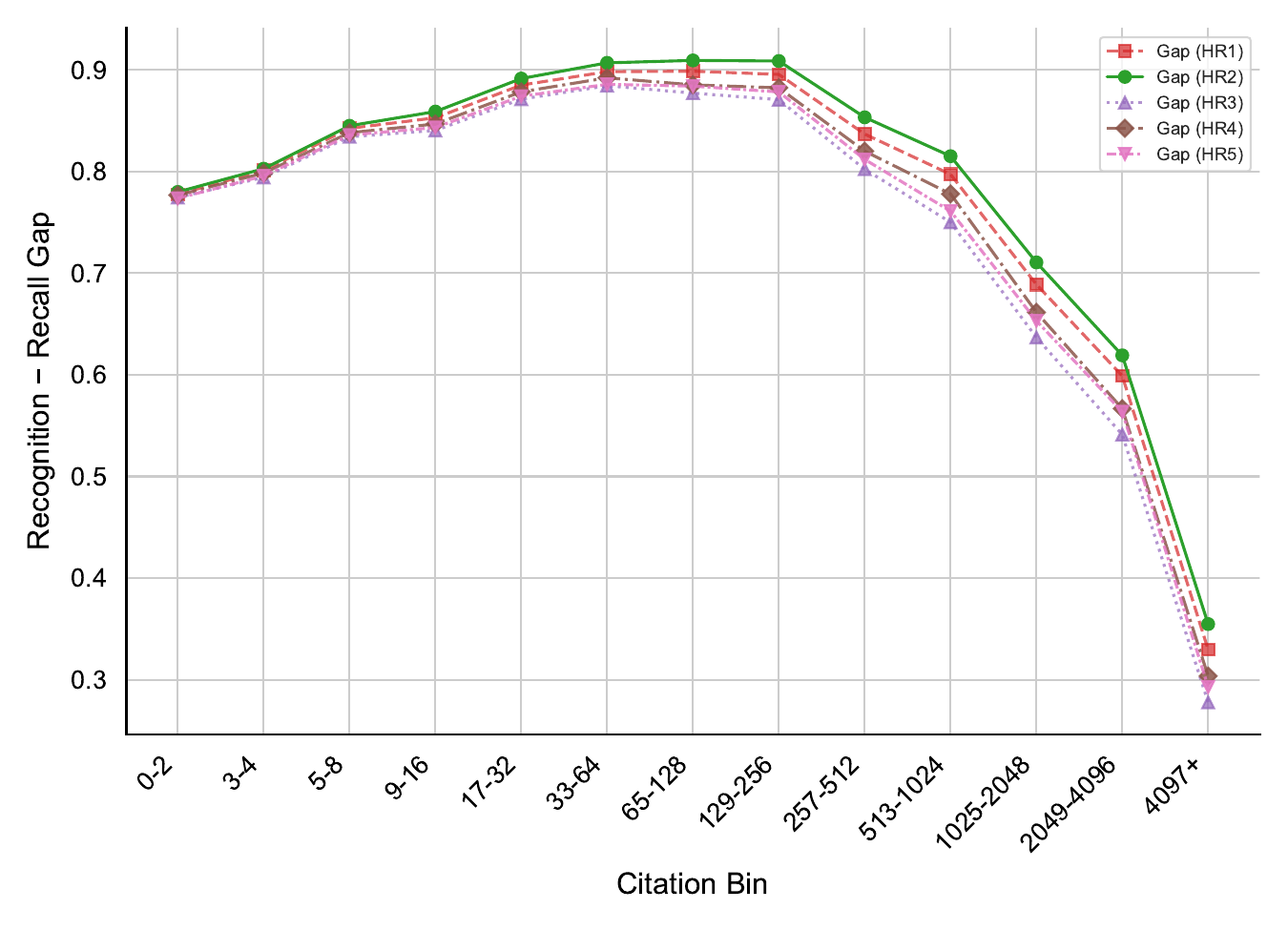}
  \caption{Recall--recognition gap computed under all five HR variants (HR$_1$--HR$_5$) across citation bins.}
  \label{fig:hr_variants_gap}
\end{figure}

\noindent\textbf{Evaluation Metrics.}
\textit{Raw MC accuracy} $\bar{a}$ is the fraction of correctly answered items. \textit{Chance-corrected accuracy} removes the contribution of random guessing (assuming a 4-option forced choice with 0.25 chance baseline):
\begin{equation}
    \tilde{a} = \frac{\bar{a} - 0.25}{0.75}
    \label{eqn:chance_corrected}
\end{equation}
\textit{Recall accuracy} is defined as $1 - \text{HR}_2$ from the GPT-4o open-ended results. These per-paper $\text{HR}_2$ values are taken from the LLM self-evaluation pipeline of Appendix~\ref{sec:appendix_eval_comparison}; the two evaluation methods agree closely (Table~\ref{tab:eval_agreement}), so the gap estimates are not sensitive to this choice. We adopt $\text{HR}_2$ as the canonical recall metric because all five HR variants ($\text{HR}_1$--$\text{HR}_5$) yield nearly identical gap estimates across all 13 citation bins (Figure~
\ref{fig:hr_variants_gap}). The \textit{recall--recognition gap} captures the additional correct identifications enabled by the structured recognition format:
\begin{equation}
    \Delta = \tilde{a} - (1 - \text{HR}_2)
    \label{eqn:recall_recog_gap}
\end{equation}
The \textit{residual recognition rate} is the fraction of papers where recall fails ($\text{HR}_2 > 0.5$) that are nonetheless correctly recognized under MC.

\subsection{Results}

Table~\ref{tab:mc_full_results} shows MC recognition accuracy, recall accuracy, and the recall--recognition gap across citation bins. Even for 0--2 citation papers, chance-corrected MC accuracy reaches 78.9\%, far above the 25\% chance baseline (one-sample $t = 45.7$, $p \approx 10^{-224}$). Recognition accuracy increases monotonically with citation count (Spearman $\rho = 0.956$, $p = 3.4 \times 10^{-7}$), reaching 100\% for the highest citation bin.

\setlength{\tabcolsep}{3pt}
\begin{table}[h!]
  \centering
  \begin{tabular}{@{}lrrr@{}}
    \toprule
    \textbf{Citation Bin} & \textbf{Recall acc.} & \textbf{MC acc. (CC)} & \textbf{Gap} \\
    \midrule
    0--2       & 0.009 & 0.789 & 0.780 \\
    3--4       & 0.022 & 0.825 & 0.803 \\
    5--8       & 0.013 & 0.858 & 0.845 \\
    9--16      & 0.034 & 0.893 & 0.859 \\
    17--32     & 0.033 & 0.925 & 0.892 \\
    33--64     & 0.038 & 0.945 & 0.907 \\
    65--128    & 0.057 & 0.967 & 0.909 \\
    129--256   & 0.075 & 0.983 & 0.909 \\
    257--512   & 0.132 & 0.985 & 0.853 \\
    513--1024  & 0.182 & 0.997 & 0.815 \\
    1025--2048 & 0.271 & 0.982 & 0.711 \\
    2049--4096 & 0.373 & 0.992 & 0.619 \\
    4097+      & 0.645 & 1.000 & 0.355 \\
    \bottomrule
  \end{tabular}
  \caption{Full recall--recognition results across all citation bins. Recall acc.\ $= 1 - \text{HR}_2$ (GPT-4o open-ended). MC acc.\ (CC) = chance-corrected MC recognition accuracy. Gap = recall--recognition gap. }
  \label{tab:mc_full_results}
  \vspace{-0.2cm}
\end{table}

The gap grows through the lower citation bins as recognition improves rapidly while recall remains near zero, then peaks at 0.909 for the 65--256 range, the point at which recognition first saturates near ceiling, yet recall is still low.
Beyond this range, the gap narrows as recall begins to catch up with rising citation counts, though a gap of 0.355 persists even at the highest bin, confirming that the disparity is robust across the full popularity spectrum.

Table~\ref{tab:residual_knowledge} summarizes residual recognition rates across recall-failure strata.
Among the 8{,}342 papers where recall substantially fails ($\text{HR}_2 > 0.5$, representing 91.6\% of the dataset), 93.9\% are nonetheless correctly recognized under MC. The residual recognition rate remains high under a stricter threshold ($\text{HR}_2 > 0.75$: 93.6\%). Together, these results indicate that the model's parametric knowledge of author teams is substantially richer than open-ended generation can surface.

\setlength{\tabcolsep}{4pt}
\begin{table}[htbp]
  \centering
  \footnotesize
  \begin{tabular}{@{}lrrr@{}}
    \toprule
    \textbf{Stratum} & \textbf{Papers} & \textbf{Correct recog.} & \textbf{Residual rate} \\
    \midrule
    $\text{HR}_2 > 0.5$ & 8{,}342& 7{,}834 & 93.9\% \\
    $\text{HR}_2 > 0.75$ & 7{,}947& 7{,}440 & 93.6\% \\
    \bottomrule
  \end{tabular}
  \caption{Residual recognition rates: papers where recall fails are nonetheless correctly recognized under MC, across two recall-failure thresholds.}
  \label{tab:residual_knowledge}
\end{table}

\subsection{Harder Distractor Controls}
\label{sec:appendix_hard_distractors}

The distractor teams in the main setup are drawn at random from the same field, so a model might select the correct option from partial cues, such as recognizing a single familiar name, rather than from knowledge of the full author list. We therefore repeated the experiment with two harder constructions, keeping the model, the 9{,}108 papers, the prompt, and the scoring unchanged.

\noindent\textbf{Same-field one-name swap.} We keep the true author list and its order and replace one randomly chosen author with a same-field author, so each of the three distractors shares $k-1$ of the $k$ true authors.

\noindent\textbf{Collaborator one-name swap.} We keep the true first author fixed and replace one of the remaining positions with a real collaborator of that first author, retrieved from the first author's other papers through the Semantic Scholar API. Each option is then a plausible team of real, subfield-matched people that differs from the truth by a single name. For single-author papers, the three distractors are three distinct collaborators of the true author. 96.7\% of first authors had at least three usable collaborators; the remainder fall back to the same-field construction.

Table~\ref{tab:mc_hard_distractors} reports chance-corrected accuracy and the recall--recognition gap under all three constructions. The same-field swap is strictly harder than the original setup, yet recognition is almost unchanged (0--2 bin: 0.789 to 0.782), so the conclusion of Section~\ref{sec:mc_recognition} is robust to distractors that share $k-1$ correct authors. The collaborator swap is the strongest control and does reduce recognition at the long tail (0--2 bin: 0.789 to 0.528), which is consistent with the concern that random distractors overstate recognition for rarely cited papers. The remaining signal is still far above the 0.25 chance baseline (one-sample $t = 23.3$, $p < 10^{-90}$), and 64.4\% of the 0--2 citation papers whose open-ended recall fails are recognized correctly. Among the 8{,}342 papers with $\text{HR}_2 > 0.5$, the residual recognition rate is 91.5\% under the same-field swap and 76.9\% under the collaborator swap, compared with 93.9\% in the main setup. The recall--recognition gap therefore holds under both harder controls, at a more conservative magnitude under the collaborator swap.

\setlength{\tabcolsep}{3pt}

\begin{table}[htbp]
  \centering
  \footnotesize
  \resizebox{\columnwidth}{!}{%
  \begin{tabular}{@{}lrrrrrrr@{}}
    \toprule
    & & \multicolumn{3}{c}{\textbf{MC acc. (CC)}} & \multicolumn{3}{c}{\textbf{Gap}} \\
    \cmidrule(lr){3-5}\cmidrule(lr){6-8}
    \textbf{Bin} & \textbf{Recall} & Orig. & S-F & Col. & Orig. & S-F & Col. \\
    \midrule
    0--2       & 0.009 & 0.789 & 0.782 & 0.528 & 0.780 & 0.773 & 0.519 \\
    3--4       & 0.022 & 0.825 & 0.818 & 0.564 & 0.803 & 0.796 & 0.543 \\
    5--8       & 0.013 & 0.858 & 0.852 & 0.582 & 0.845 & 0.838 & 0.568 \\
    9--16      & 0.034 & 0.893 & 0.863 & 0.613 & 0.859 & 0.829 & 0.579 \\
    17--32     & 0.033 & 0.925 & 0.898 & 0.655 & 0.892 & 0.865 & 0.622 \\
    33--64     & 0.038 & 0.945 & 0.903 & 0.743 & 0.907 & 0.865 & 0.705 \\
    65--128    & 0.057 & 0.967 & 0.915 & 0.727 & 0.909 & 0.858 & 0.669 \\
    129--256   & 0.075 & 0.983 & 0.940 & 0.787 & 0.909 & 0.865 & 0.712 \\
    257--512   & 0.132 & 0.985 & 0.952 & 0.823 & 0.853 & 0.820 & 0.692 \\
    513--1024  & 0.182 & 0.997 & 0.943 & 0.818 & 0.815 & 0.762 & 0.637 \\
    1025--2048 & 0.271 & 0.982 & 0.940 & 0.868 & 0.711 & 0.669 & 0.596 \\
    2049--4096 & 0.373 & 0.992 & 0.939 & 0.865 & 0.619 & 0.566 & 0.492 \\
    4097+      & 0.645 & 1.000 & 0.934 & 0.945 & 0.355 & 0.289 & 0.300 \\
    \bottomrule
  \end{tabular}%
  }
  \caption{Recognition under the original distractors (Orig.), the same-field one-name swap (S-F), and the collaborator one-name swap (Col.). Recall $= 1 - \text{HR}_2$; MC acc.\ (CC) is chance-corrected; Gap $=$ MC acc.\ (CC) $-$ recall.}
  \label{tab:mc_hard_distractors}
\end{table}

\subsection{Binary Perfect-Recall}
\label{sec:appendix_binary_recall}

A potential concern is that $1-\text{HR}_2$ (partial-credit, Jaccard-based) and MC accuracy (binary correct/incorrect) are not directly comparable. To address this, we additionally compute a \textit{binary perfect-recall} metric: a paper is counted as correctly recalled only if the model produces all ground-truth authors with no hallucinations and no omissions ($H = 0$ and $U = 0$). This is a fully binary metric, directly comparable to MC accuracy.

\begin{figure}[htbp]
  \centering
  \includegraphics[width=\columnwidth]{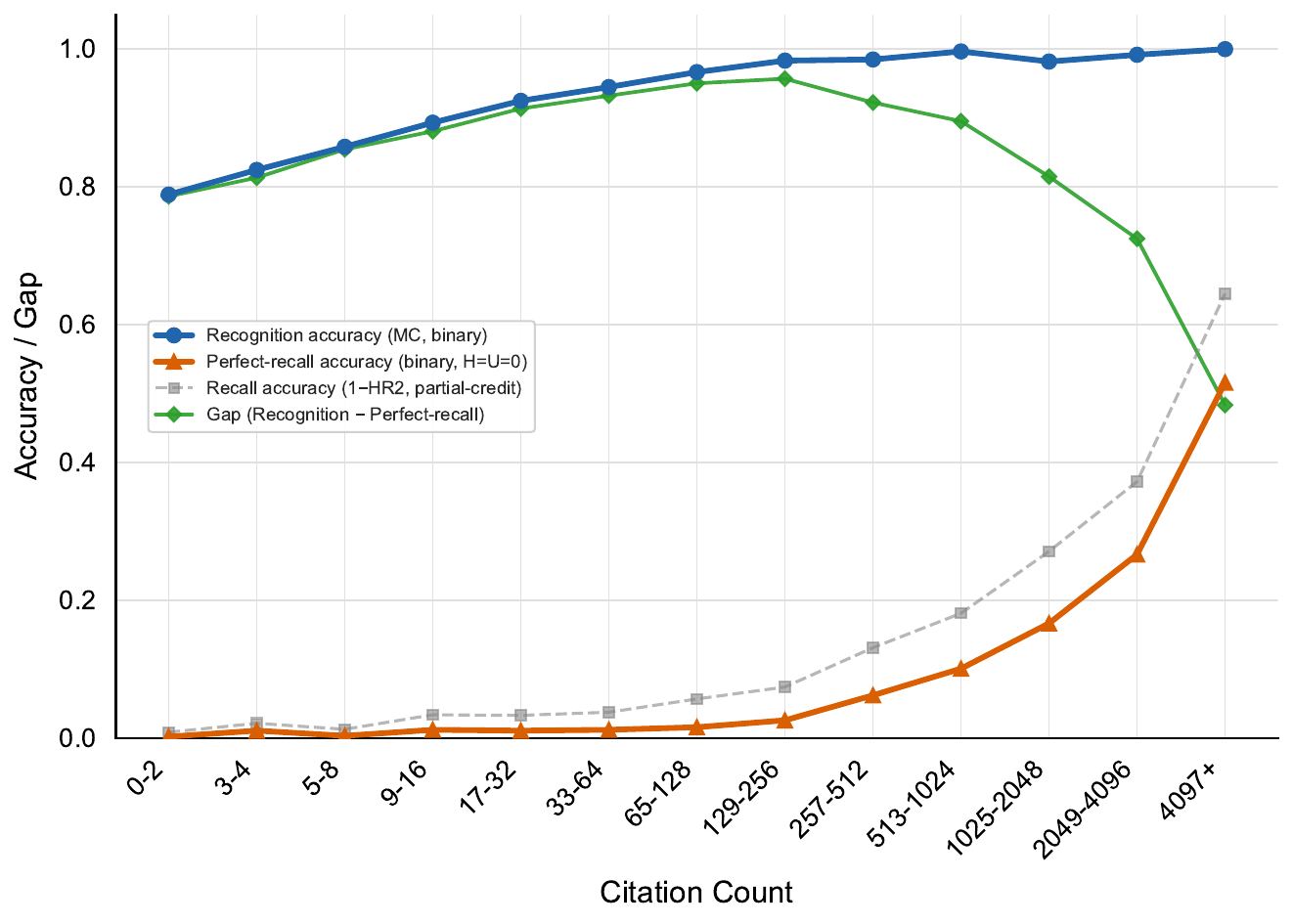}
  \caption{Binary perfect-recall accuracy ($H$=$U$=0, no partial credit) vs.\ chance-corrected MC recognition accuracy across citation bins. The grey dashed line shows the partial-credit $1-\text{HR}_2$ recall for reference.}
  \label{fig:perfect_recall_comparison}
\end{figure}

Figure~\ref{fig:perfect_recall_comparison} shows that the recall--recognition gap is even larger under this stricter criterion: binary perfect-recall is substantially lower than $1-\text{HR}_2$ at every citation bin, while MC accuracy is unchanged. The gap under the binary metric exceeds 0.78 in every bin below 512 citations and peaks at 0.957 in the 129--256 bin. This confirms that the recall--recognition gap is not an artifact of partial credit in the recall metric.

\section{More Details on Mechanistic Evidence for Prevalence Encoding}
\label{sec:appendix_prevalence_encoding}

The main paper shows that hallucination rates decrease as citation count increases.
In this section, we ask whether this prevalence signal is also visible in
the internal states of open-source language models.  This analysis is intended as
technical support for the prevalence hypothesis.  We evaluate Qwen3-32B and Mistral-Small-3.2-24B on the same
9,108 papers used in Section~\ref{sec:methodology}.

We extract hidden states at several query- and entity-level positions. P1 is the last prompt token, before any answer token is generated; P2 is the first generated token; P3 is the last generated token; P4 is the mean-pooled prompt state; and P5 is the mean-pooled generated-answer state.  We also extract P9
states at the last token of each generated author name for entity-level
analyses.

\subsection{The Prevalence Signal is Strongest Before Generation}

\begin{figure}[t]
\centering
\includegraphics[width=\columnwidth]{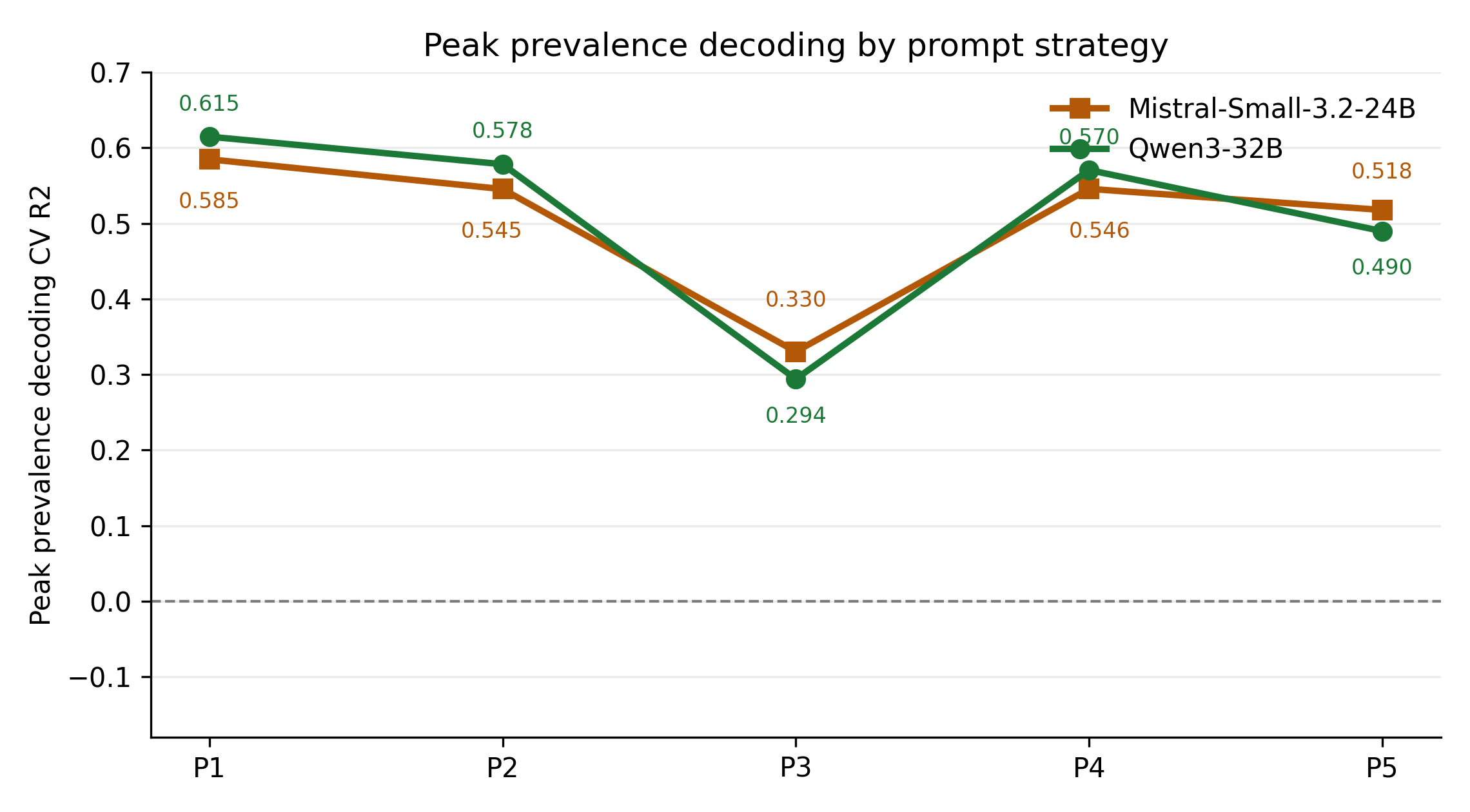}
\caption{Peak prevalence decoding performance across extraction strategies.
P1, the last prompt token, is strongest for both open-source models.}
\label{fig:appendix_strategy_comparison}
\end{figure}

\begin{figure}[t]
\centering
\includegraphics[width=\columnwidth]{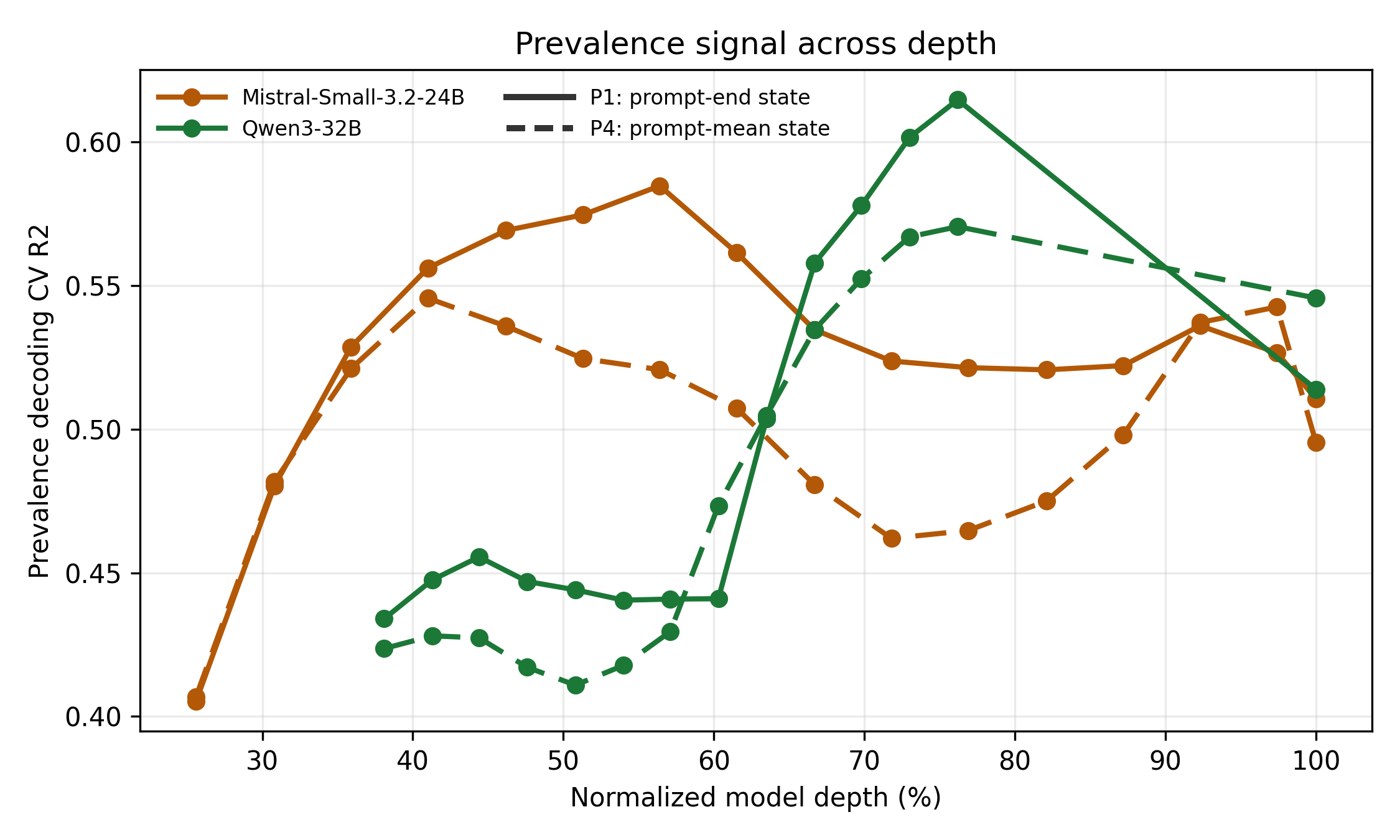}
\caption{Layer-wise prevalence decoding from P1 and P4 representations.  The
P1 prevalence signal strengthens through depth and peaks in mid-to-late layers.}
\label{fig:appendix_prevalence_depth}
\end{figure}

We first compare the extraction strategies P1--P5 across layers
(Figure~\ref{fig:appendix_strategy_comparison}). \textbf{Prevalence signal} refers to how accurately a linear decoder (ridge regression) fitted to the hidden state can predict $\log(\text{citations}+1)$, measured by cross-validated $R^2$. We find that P1 shows the
strongest prevalence signal for both models, while P3, the last generated-token state, is weak.  This pattern suggests that the prevalence signal is already
present at the end of prompt processing, before the model begins generating the
author list, rather than being merely a post-generation artifact.  The layer-wise
trajectory in Figure~\ref{fig:appendix_prevalence_depth} further shows that this
P1 signal strengthens through model depth and peaks in mid-to-late layers.
Because P1 captures the model state before answer generation and shows the
strongest prevalence signal in our experiments, the remaining technical analyses focus on P1 unless otherwise stated.

\subsection{Internal States Predict Paper Prevalence}

We next test whether a paper's citation prevalence can be decoded from the P1
state, following standard probing setups for measuring information recoverable from model representations~\citep{hewitt2019structural, tenney2019bert}.  For each model, we use the P1 hidden state at the strongest layer (layer 48 for Qwen3-32B and layer 22 for Mistral-Small-3.2-24B), reduce the embedding to 100 SVD components, and fit a ridge regression to predict $\log(\mathrm{citations}+1)$ under cross-validation.  The results show that the model's pre-generation representation contains a strong signal of paper prevalence (Table~\ref{tab:appendix_prevalence_decoding}).
We define the P1-decoded prevalence score as the out-of-fold prediction of
$\log(\mathrm{citations}+1)$ produced by this ridge decoder; thus, each paper's score is produced by a decoder that was not trained on that paper.

\setlength{\tabcolsep}{4pt}
\begin{table}[t]
\centering
\footnotesize

\begin{tabular}{lrrrr}
\toprule
\textbf{Model} & \textbf{Layer} & \textbf{Depth} & \textbf{CV $R^2$} & \textbf{Spearman $\rho$} \\
\midrule
Qwen3-32B & 48 & 76.2\% & 0.615 & 0.776 \\
Mistral-24B & 22 & 56.4\% & 0.585 & 0.759 \\
\bottomrule
\end{tabular}
\caption{P1 hidden states predict paper prevalence. The target is
$\log(\mathrm{citations}+1)$, and the representation is reduced to 100 SVD
components before ridge regression.}
\label{tab:appendix_prevalence_decoding}
\end{table}

This result is not explained by simple metadata controls, and the within-field analysis further shows that it is not driven only by coarse disciplinary identity.  Metadata controls alone explain only
3.9\% of the variance in log citations, whereas the P1 embedding explains
substantially more.  Adding P1 embeddings on top of the metadata controls
increases cross-validated $R^2$ to 0.620 for Qwen3-32B and 0.590 for
Mistral-Small-3.2-24B (Table~\ref{tab:appendix_adjusted_decoding}).

\setlength{\tabcolsep}{4pt}
\begin{table}[t]
\centering
\footnotesize

\begin{tabular}{llrr}
\toprule
\textbf{Model} & \textbf{Feature set} & \textbf{CV $R^2$} & \textbf{Spearman $\rho$} \\
\midrule
Qwen3-32B & Metadata controls & 0.039 & 0.180 \\
Qwen3-32B & P1 embedding & 0.615 & 0.776 \\
Qwen3-32B & Metadata + P1 & 0.620 & 0.780 \\
\midrule
Mistral-24B & Metadata controls & 0.039 & 0.180 \\
Mistral-24B & P1 embedding & 0.585 & 0.759 \\
Mistral-24B & Metadata + P1 & 0.590 & 0.764 \\
\bottomrule
\end{tabular}
\caption{Prevalence decoding after accounting for simple metadata controls.  The
metadata controls include field, year, author count, and title length.}
\label{tab:appendix_adjusted_decoding}
\end{table}

As a stronger check against coarse field confounding, we repeated the prevalence-decoding analysis separately within each academic field.
In this within-field setting, field identity is held fixed by construction, so the decoder must predict variation in $
\log(\mathrm{citations}+1)$ among papers from the same discipline. P1 embeddings remained strongly predictive: across the eight fields,
Qwen3-32B achieved mean within-field CV $R^2=0.589$ and mean Spearman $\rho=0.760$, while Mistral-Small-3.2-24B achieved mean within-field
CV $R^2=0.563$ and mean Spearman $\rho=0.746$. By comparison, metadata controls without field identity explained little within-field
variation (mean CV $R^2=0.040$), reducing the concern that the P1 signal is driven only by coarse disciplinary identity. A finer-grained lexical or topical confound could still contribute, which we test next.

We therefore fit a title-only baseline that predicts $\log(\mathrm{citations}+1)$ from the title text alone (TF-IDF and SBERT features, ridge regression) and residualize the P1 probe on it, using the same fold structure as the main probe. The title-only baseline reaches CV $R^2 = 0.13$ for both models. After removing everything it explains, P1 still attains CV $R^2 = 0.520$ for Qwen3-32B and 0.483 for Mistral-Small-3.2-24B. The prevalence signal in P1 is therefore not recoverable from the title string, which is consistent with paper-specific familiarity rather than topic popularity.

\subsection{Negative Controls}

To rule out high-dimensional regression artifacts, we repeat the same decoding
pipeline after randomly shuffling citation labels.  The shuffled-label controls
collapse to near-zero or negative cross-validated $R^2$ for both models, while
the real-label P1 decoding remains strong (Table~\ref{tab:appendix_negative_control}).

\setlength{\tabcolsep}{1pt}
\begin{table}[t]
\centering
\footnotesize
\resizebox{\columnwidth}{!}{%
\begin{tabular}{lrr}
\toprule
\textbf{Model} & \textbf{Real-label CV $R^2$} & \textbf{Shuffled-label mean CV $R^2$} \\
\midrule
Qwen3-32B & 0.615 & $-0.003$ \\
Mistral-24B & 0.585 & $-0.002$ \\
\bottomrule
\end{tabular}%
}
\caption{Negative controls for prevalence decoding. Shuffling citation labels
removes the signal, indicating that prevalence decoding is not a generic artifact
of fitting high-dimensional embeddings.}
\label{tab:appendix_negative_control}
\end{table}

\subsection{Linear Decodability Across Citation Regimes}

To make the prevalence-decoding result more interpretable, we also evaluate
classification versions of the same question.  Instead of predicting the exact
log citation count, these probes ask whether the P1 representation can separate
coarser prevalence regimes, such as long-tail papers versus highly cited papers.
For the binary tasks, we use logistic regression with cross-validated out-of-fold predictions.  We evaluate four binary tasks: \textit{zero-versus-cited}
separates papers with zero citations from papers with at least one citation;
\textit{above-median} separates papers above versus below the median citation count;
\textit{head-versus-tail} compares the bottom and top citation quartiles; and \textit{top-decile}
separates the top 10\% most-cited papers from the rest.
The head-versus-tail task is especially strong: Qwen3-32B achieves AUROC 0.975 and Mistral-Small-3.2-24B achieves AUROC 0.970. This strong performance is expected: by excluding the ambiguous middle of the citation distribution and comparing only the two extremes, the prevalence signal is at its cleanest, making the classification boundary easiest to learn from the P1 representation.
The above-median and top-decile probes are also strong, with AUROC 0.895/0.888
and 0.946/0.930 for Qwen3-32B and Mistral-Small-3.2-24B, respectively (Table~\ref{tab:appendix_bucket_probe}).

\setlength{\tabcolsep}{4pt}
\begin{table}[t]
\centering
\footnotesize
\resizebox{\columnwidth}{!}{%
\begin{tabular}{llrr}
\toprule
\textbf{Model} & \textbf{Task} & \textbf{AUROC} & \textbf{Balanced acc.} \\
\midrule
Qwen3-32B & Zero vs. cited & 0.833 & 0.741 \\
Qwen3-32B & Above median & 0.895 & 0.812 \\
Qwen3-32B & Head vs. tail & \textbf{0.975} & \textbf{0.920} \\
Qwen3-32B & Top decile & 0.946 & 0.873 \\
\midrule
Mistral-24B & Zero vs. cited & 0.832 & 0.746 \\
Mistral-24B & Above median & 0.888 & 0.804 \\
Mistral-24B & Head vs. tail & \textbf{0.970} & \textbf{0.911} \\
Mistral-24B & Top decile & 0.930 & 0.863 \\
\bottomrule
\end{tabular}%
}
\caption{Citation-prevalence classification from P1 hidden states. The
head-versus-tail task compares papers below the first citation quartile with
papers above the third citation quartile.}
\label{tab:appendix_bucket_probe}
\end{table}

\subsection{Papers the Model Represents as More Prevalent Have Lower Error}

We connect the internal prevalence signal back to the main paper's
hallucination result.  For each paper, the P1 decoder produces an out-of-fold
estimate of $\log(\mathrm{citations}+1)$ using only the model's hidden state.
Intuitively, this score measures how much prevalence information is present in
the model's hidden state.  We then sort papers by this P1-decoded prevalence
score and split them into ten equal-sized groups (deciles), where decile 10
contains the top 10\% highest-scoring papers.  Hallucination rates decrease from
the lowest-scoring decile to the highest-scoring decile.
For Qwen3-32B, mean HR$_2$ falls from 0.994 in the lowest group to 0.843 in the
highest group.  For Mistral-Small-3.2-24B, mean HR$_2$ falls from 0.995 to 0.879
(Table~\ref{tab:appendix_internal_deciles}).

\setlength{\tabcolsep}{2pt}
\begin{table}[t]
\centering

\resizebox{\columnwidth}{!}{%
\begin{tabular}{lrrr}
\toprule
\textbf{Model} & \textbf{P1-decoded prevalence decile} & \textbf{Mean citations} & \textbf{Mean HR$_2$} \\
\midrule
Qwen3-32B & 1 (lowest) & 13.4 & 0.994 \\
Qwen3-32B & 10 (highest) & 2393.6 & 0.843 \\
\midrule
Mistral-24B & 1 (lowest) & 12.6 & 0.995 \\
Mistral-24B & 10 (highest) & 2230.5 & 0.879 \\
\bottomrule
\end{tabular}%
}
\caption{Hallucination rate by P1-decoded prevalence decile.}
\label{tab:appendix_internal_deciles}
\end{table}

\subsection{Mediation Analysis: Hidden-State Prevalence and Hallucination}

The preceding analyses show that P1 hidden states encode paper prevalence and
that papers with higher P1-decoded prevalence tend to have lower error rates.   We therefore use a modern mediation analysis framework~\citep{imai2010general, tingley2014mediation} to quantify how much of the citation--hallucination association is carried by the internal prevalence signal. The treatment is a continuous contrast in $\log(\mathrm{citations}+1)$ from the empirical 25th to 75th percentile, the mediator is the out-of-fold prevalence estimate decoded from P1 embeddings, and the outcome is HR$_2$. We estimate the indirect path through the mediator, the remaining direct association, and the mediated proportion using bootstrap confidence intervals. The P1 prevalence mediator accounts for 67.3\% of the citation--hallucination effect for
Qwen3-32B and 51.9\% for Mistral-Small-3.2-24B (Table~\ref{tab:appendix_mediation_summary}). This shows that the prevalence signal decoded from pre-generation representations is strongly aligned with hallucination risk: papers represented as more prevalent tend to elicit fewer hallucinated authors~(Table~\ref{tab:appendix_mediation_summary}).

\begin{table}[t]
\centering

\resizebox{\columnwidth}{!}{%
\begin{tabular}{lrrr}
\toprule
\textbf{Model} & \textbf{P1 layer} & \textbf{HR$_2$ mediated} & \textbf{Mediated proportion 95\% CI} \\
\midrule
Qwen3-32B & 48 & 67.3\% & $[57.3\%, 76.7\%]$ \\
Mistral-24B & 22 & 51.9\% & $[43.3\%, 61.0\%]$ \\
\bottomrule
\end{tabular}%
}
\caption{Mediation analysis using the P1-decoded internal prevalence signal. The mediator is the out-of-fold predicted prevalence score
decoded from the model's pre-generation hidden representation, and the outcome is HR$_2$.}
\label{tab:appendix_mediation_summary}
\end{table}

\subsection{Activation Steering on the Prevalence Direction}
\label{sec:appendix_steering}

The analyses above are correlational. To test whether the pre-generation prevalence signal has any influence on attribution behavior, we intervene on it directly. At the peak-decodability layer (layer 48 for Qwen3-32B and layer 22 for Mistral-Small-3.2-24B), we construct a steering vector as the difference in mean hidden states between correctly attributed and hallucinated papers, following the difference-of-means approach~\citep{rimsky-etal-2024-steering}. We then add this vector with a negative coefficient ($\alpha = -2$) while the model answers papers it otherwise attributes correctly ($n = 50$ per model), and compare against five norm-matched random directions.

Steering against this direction sharply increases hallucination: HR$_2$ rises from 10\% to 70\% for Qwen3-32B and from 36\% to 82\% for Mistral-Small-3.2-24B (Table~\ref{tab:appendix_steering}), increases of 60 points (95\% CI 44--74, McNemar $p < 10^{-7}$) and 46 points (95\% CI 32--60, $p < 10^{-5}$). Both exceed the largest effect of any random direction (at most 48\% and 52\%). Repeating the intervention along the prevalence probe direction, which is defined by citation count rather than by the hallucination outcome, reproduces the effect (58\% and 64\%).

The effect is asymmetric: steering toward the prevalence direction does not restore recall on papers the model fails, so we do not claim that hallucination can be repaired this way, nor that the decoded direction is the mechanism by which these errors arise. The sample is also small ($n = 50$ per model). We read the result as convergent evidence that prevalence is tracked internally in a form that can shift attribution behavior when perturbed.

\setlength{\tabcolsep}{4pt}
\begin{table}[t]
\centering
\footnotesize
\resizebox{\columnwidth}{!}{%
\begin{tabular}{lrrr}
\toprule
\textbf{Model} & \textbf{Baseline} & \textbf{Mean-difference steering} & \textbf{Random directions} \\
\midrule
Qwen3-32B & 10\% & 70\% & 30--48\% \\
Mistral-24B & 36\% & 82\% & 48--52\% \\
\bottomrule
\end{tabular}%
}
\caption{Hallucination rate (HR$_2$) under activation steering at $\alpha = -2$, with $n = 50$ papers per model that the model attributes correctly without intervention. The last column reports the range over five norm-matched random directions.}
\label{tab:appendix_steering}
\end{table}

\subsection{Entity-Level Analysis with P9 Name States}

The P1 analyses above operate at the query level. As an entity-level supplement, we also analyze P9 hidden states, where each P9 vector is extracted at the last token of a generated author name. Thus, if a model outputs five author names, the run contributes five P9 vectors for that paper. Each generated name is then labeled as matched or hallucinated using the same author-matching pipeline used for the main hallucination metrics.

Matched and hallucinated names show moderate separability in this P9 space. A class-balanced logistic probe distinguishes matched from hallucinated generated names with AUROC 0.814 for Qwen3-32B and 0.818 for Mistral-Small-3.2-24B.

Because hallucination-generated names dominate the entity-level dataset (97.3\% for Qwen3-32B and 97.5\% for Mistral-Small-3.2-24B), we also report imbalance-aware and probe-free checks. AUPRC lift is only 1.019 for Qwen and 1.018 for Mistral, indicating that average precision is only slightly above the prevalence baseline. The normalized centroid-separation check gives 0.246 for Qwen and 0.209 for Mistral, with nearest-centroid balanced accuracy of 0.760 and 0.743, respectively (Table~\ref{tab:appendix_p9_entity}).

Thus, P9 provides supporting entity-level evidence that recalled and guessed names occupy partially different representation regions, but the effect is modest and secondary to the P1 prevalence-encoding results.

\begin{table}[t]
\centering
\captionsetup{font=small}
\footnotesize

\resizebox{\columnwidth}{!}{%
\begin{tabular}{lrrrrrr}
\toprule
\textbf{Model} & \textbf{Matched} & \textbf{Halluc.} & \textbf{AUROC} & \textbf{AUPRC lift} & \textbf{Norm. centroid} & \textbf{NC bal. acc.} \\
\midrule
Qwen3-32B & 1,286 & 46,726 & 0.814 & 1.019 & 0.246 & 0.760 \\
Mistral-24B & 1,303 & 49,960 & 0.818 & 1.018 & 0.209 & 0.743 \\
\bottomrule
\end{tabular}%
}
\caption{Entity-level P9 analysis. Each row is a generated author name labeled as matched or hallucinated. AUPRC lift adjusts the average precision for the class-imbalance baseline; Norm. centroid and NC bal. acc. summarize centroid-based separation.}
\label{tab:appendix_p9_entity}
\end{table}

\section{Additional Discussion}
\label{sec:discussion}

\subsection{The Long Tail of Factual Recall}

Citation count serves as a practical proxy for training-data presence: highly cited papers are reinforced across academic databases, syllabi, Wikipedia, and downstream citations. The magnitude of this effect is substantial: hallucination rates in GPT-4o drop by nearly 63 percentage points between the lowest and highest citation bins. Yet even the most famous papers (top 1\% by citations) retain a 36.4\% error rate, revealing that frequency-based memorization cannot guarantee reliable recall of structured relational facts (Author $\leftrightarrow$ Paper). This residual failure ceiling indicates that retrieval augmentation is not merely helpful but necessary for production bibliographic systems.

\subsection{Domain-Specific ``Digital Footprints''}

The variation in correlation strength across fields offers insight into corpus composition. Computer Science displayed the strongest correlation ($\rho = -0.476$, GPT-4o), likely due to the field's unique open-access culture (arXiv), high digitization, and extensive discussion on crawled platforms (GitHub, StackOverflow, tech blogs). This creates a stronger signal-to-noise ratio for CS metadata in the training set.

In contrast, Materials Science showed the weakest response to citation impact ($\rho = -0.228$, GPT-4o). We hypothesize this stems from a smaller "digital footprint". Many materials science papers may reside behind stricter paywalls or in PDF-only formats that are harder to parse during training data curation.

Critically, we observed a decoupling of complexity and error. Medicine (highest mean author count: 5.0) performed better than Materials Science (4.2 authors), indicating that task difficulty is determined not by answer length but by the familiarity of the entities involved.

\subsection{Consistency Across Architectures}

Beyond field-level differences, a cross-model comparison reveals a counterintuitive scaling pattern: the newer models (Claude Sonnet 4.5 and DeepSeek-R1) exhibit stronger negative correlations with citation count than GPT-4o. This implies that increased model capability amplifies sensitivity to training-data frequency, consolidating performance gains on head-of-distribution knowledge while leaving long-tail recall equally vulnerable. These findings challenge the assumption that scaling alone will resolve the hallucination problem for low-prevalence facts.

The hidden-state analyses on Qwen3-32B and Mistral-Small-3.2-24B provide a complementary open-source check: although these are not the same systems as the closed-source behavioral models, the same prevalence signal is encoded in their accessible internal representations.

\subsection{From Retrieval Failure to Latent Knowledge: Implications for Mitigation}

The recognition experiment (Section~\ref{sec:mc_recognition}) and hidden-state
probing results (Section~\ref{sec:prevalence_encoding}) reframe
prevalence-driven hallucination as an access bottleneck rather than only knowledge absence. The recognition experiment shows that for 93.9\% of papers where
open-ended recall substantially fails, the model nonetheless selects the correct
author team under multiple-choice constraints, suggesting that relevant factual
information is often available in a form that open-ended generation does not
reliably surface. The hidden-state analysis further shows that paper prevalence
is encoded in pre-generation representations, and mediation analysis indicates that this internal prevalence signal accounts for a large share of the
citation--hallucination association. Together, these findings suggest that low-prevalence papers are not simply absent from model knowledge; rather, their author information is often difficult to access reliably during generation.

This reframing has practical implications for mitigation. Because the bottleneck
often appears to be retrieval rather than memorization alone, approaches that
surface latent knowledge, such as reasoning or verification scaffolds before
committing to an answer~\citep{wei2022chain}, constrained decoding guided by
entity-state separability, or activation-space interventions on the P1 prevalence
representation~\citep{li2023inference, zou2023representation}, may be more
tractable than simply augmenting training data. For bibliographic systems that must produce verifiable attributions,
relying only on internal knowledge is therefore unsafe;
Retrieval-Augmented Generation (RAG) is likely necessary for reliable
performance in this domain. RAG addresses the bottleneck from the outside by
supplying verified context, while future mechanistic approaches could target it
from within.

\end{document}